\documentclass[dvipsnames]{article} %
\usepackage{template_conference}

\usepackage{booktabs}
\usepackage{graphicx}
\usepackage{enumitem}
\usepackage{wrapfig}
\usepackage{algorithm}
\usepackage{algpseudocode}
\usepackage[utf8]{inputenc}
\usepackage[T1]{fontenc}
\DeclareUnicodeCharacter{FF5E}{\textasciitilde}

\usepackage{adjustbox}
\usepackage{amsmath}
\usepackage{colortbl}
\usepackage[utf8]{inputenc}
\definecolor{lightgray}{rgb}{0.9,0.9,0.9}
\usepackage{caption}
\usepackage{subcaption}
\usepackage[svgnames]{xcolor}
\usepackage{setspace}
\usepackage{url}
\usepackage{multirow}
\usepackage{colortbl}
\usepackage{tabularx}
\usepackage{blindtext}
\usepackage{pgfplots}
\pgfplotsset{compat=1.18} 
\usepackage{tikz}
\usetikzlibrary{er,positioning}
\usepackage{makecell}
\usepackage{tipa}
\tipasafemode 
\usepackage{siunitx}
\usepackage{tocloft}
\usepackage{listings}
\usepackage{adjustbox}
\usepackage{xurl}
\usepackage{rotating}
\usepackage[normalem]{ulem}
\usepackage{multicol}
\usepackage{titlesec}
\usepackage{tcolorbox}
\tcbuselibrary{breakable,theorems,skins,listings}
\usepackage{transparent}
\usepackage{mathtools}

\usepackage{amsthm}

\usepackage{amsfonts}
\usepackage{etoc}
\usepackage{color}
\usepackage{autobreak}
\usepackage{soul}
\usepackage{amssymb}
\usepackage{bbding}

\newtcolorbox{codeblock}{
  colback=gray!10,   
  colframe=gray!50,  
  boxrule=0.5mm,     
  arc=2mm,           
  left=5pt,          
  top=5pt,          
  bottom=5pt,        
}

\newtcbtheorem{proposition}{Proposition}{
  fonttitle=\bfseries,
  fontupper=\normalfont,
  boxrule=0.5pt,
  left=3mm,      
  right=4mm,
  top=2mm,
  bottom=2mm,
  clip upper=false, 
  breakable=false,
  before upper=\setlength{\parindent}{1em}
}{prop}

\newtcbtheorem{definition}{Definition}{
  enhanced,
  colback=cyan!5!white,           
  colframe=cyan!45!blue!60,       
  fonttitle=\bfseries,            
  fontupper=\normalfont,        
  arc=4mm,                        
  boxrule=1pt,
  drop shadow southwest,
  attach boxed title to top left={xshift=6mm, yshift=-2mm},
  boxed title style={
    colback=cyan,
    size=small,
    arc=2mm,                      
    boxrule=0.8pt,
    left=2mm, right=2mm
  },
  before skip=10pt,
  after skip=10pt,
  left=5mm,
  right=5mm,
  top=3mm,
  bottom=3mm
}{cor}

\useunder{\uline}{\ul}{}

\usepackage{amsmath,amsfonts,bm}

\def\eqref#1{equation~\ref{#1}}

\def\1{\bm{1}}

\DeclareMathAlphabet{\mathsfit}{\encodingdefault}{\sfdefault}{m}{sl}
\SetMathAlphabet{\mathsfit}{bold}{\encodingdefault}{\sfdefault}{bx}{n}

\renewcommand{\ghlink}{https://github.com/yczhou001/PF-OPSD}

\newcommand*\justify{
  \fontdimen2\font=0.4em
  \fontdimen3\font=0.2em
  \fontdimen4\font=0.1em
  \fontdimen7\font=0.1em
  \hyphenchar\font=`\-
}

\renewcommand{\texttt}[1]{%
  \begingroup
  \ttfamily
  \begingroup\lccode`~=`/\lowercase{\endgroup\def~}{/\discretionary{}{}{}}%
  \begingroup\lccode`~=`[\lowercase{\endgroup\def~}{[\discretionary{}{}{}}%
  \begingroup\lccode`~=`.\lowercase{\endgroup\def~}{.\discretionary{}{}{}}%
  \catcode`/=\active\catcode`[=\active\catcode`.=\active
  \justify\scantokens{#1\noexpand}%
  \endgroup
}

\definecolor{blueviolet}{RGB}{138,43,226}
\newtcolorbox{insightblock}{
  colback=blueviolet!5,   
  colframe=blueviolet!50!black!50!,    
  boxrule=0.5mm,     
  arc=2mm,             
  left=0pt,        
  right=8pt,         
  top=8pt,            
  bottom=8pt,         
}

\newtcolorbox{abstractbox}{
    colback=blue!5!white,    
    frame empty,             
    boxrule=1pt,             
    arc=4mm,                
    left=8pt,                 
    right=8pt,              
    top=8pt,                  
    bottom=8pt,               
    opacityback=0.9
}

\newtcolorbox{examplebox}[1][]{
    breakable,
    colback=black!4,
    colframe=black!35,
    boxrule=0pt,
    sharp corners,
    boxsep=0pt,
    left=8pt,
    right=7pt,
    top=6pt,
    bottom=6pt,
    before upper={
        \if\relax\detokenize{#1}\relax
        \else
            {\small\bfseries\color{black!85} #1\par}
            \vspace{3pt}
        \fi
    },
    before skip=0.8\baselineskip,
    after skip=0.8\baselineskip,
    pad at break*=1.2mm
}

\title{World Models Meet Language Models: \\
On the Complementarity of Concrete and Abstract Reasoning}

\author{
\footnotesize
\textbf{Yucheng Zhou}\textsuperscript{1} \quad
\textbf{Wei Tao}\textsuperscript{2} \quad
\textbf{Yiwen Guo}\textsuperscript{3}\footnotemark[1] \quad
\textbf{Jianbing Shen}\textsuperscript{1}\footnotemark[1] \quad
\\[2pt]
\scriptsize \textsuperscript{1}University of Macau \quad
\textsuperscript{2}LIGHTSPEED \quad
\textsuperscript{3}Independent Researcher
\\[2pt]
\scriptsize \texttt{yucheng.zhou@connect.um.edu.mo}
}

\begin{document}

\maketitle
\begingroup
\renewcommand{\thefootnote}{\fnsymbol{footnote}}
\footnotetext[1]{Corresponding authors.}
\endgroup

\bibliographystyle{template_conference}
\begin{abstractbox}
\begin{center}
\textbf{\Large Abstract}
\end{center}
World models and multimodal large language models (MLLMs) provide complementary capabilities for predicting future outcomes from static visual observations. World models can generate concrete visual rollouts of possible futures, while MLLMs can reason abstractly over questions, goals, and rules. However, generated rollouts are stochastic and may be visually plausible but task-incorrect, making it necessary to determine when visual simulation is useful, whether a rollout is credible, and how it should influence the final answer. We formulate this problem as controlled concrete reasoning, where a model learns to invoke, verify, and integrate visual future simulation alongside abstract reasoning. To study this setting, we construct two human-verified benchmarks, \texttt{VRQABench} for controllable spatial lookahead and \texttt{OpenWorldQA} for open-domain physical prediction, and propose Privileged-Future On-Policy Self-Distillation (PF-OPSD). During training, PF-OPSD uses ground-truth future videos and answers only as teacher-side privileged context to evaluate on-policy concrete-reasoning trajectories, while the deployable student never observes true futures at test time. Experimental results show that PF-OPSD outperforms the baseline by 10.5 and 10.8 percentage points on \texttt{VRQABench} and \texttt{OpenWorldQA}, respectively, while increasing robustness to noisy or conflicting rollouts. Our code and dataset are available at \url{https://github.com/yczhou001/PF-OPSD}.
\end{abstractbox}
\vspace{4pt}

\section{Introduction}

Future-oriented visual reasoning asks a model to answer questions about outcomes that are not yet visible in a static observation. A single image or anchor frame may reveal objects, contacts, spatial constraints, or a puzzle state, but the answer often depends on extrapolating short-horizon dynamics or plans. Recent MLLMs can organize goals, rules, and alternatives in language \citep{SurveyMLLM,VisualMLLM,qwen3vl}, while video world models can make possible futures visually explicit through generated rollouts \citep{wan2025,HunyuanVideo,yuan2026helios,Simulatingwm,Physicalwm}. This makes future prediction a natural testbed for combining abstract language-level reasoning with concrete visual simulation.

\begin{wrapfigure}{r}{0.50\linewidth}
\centering
\vspace{-12pt}
\includegraphics[width=\linewidth]{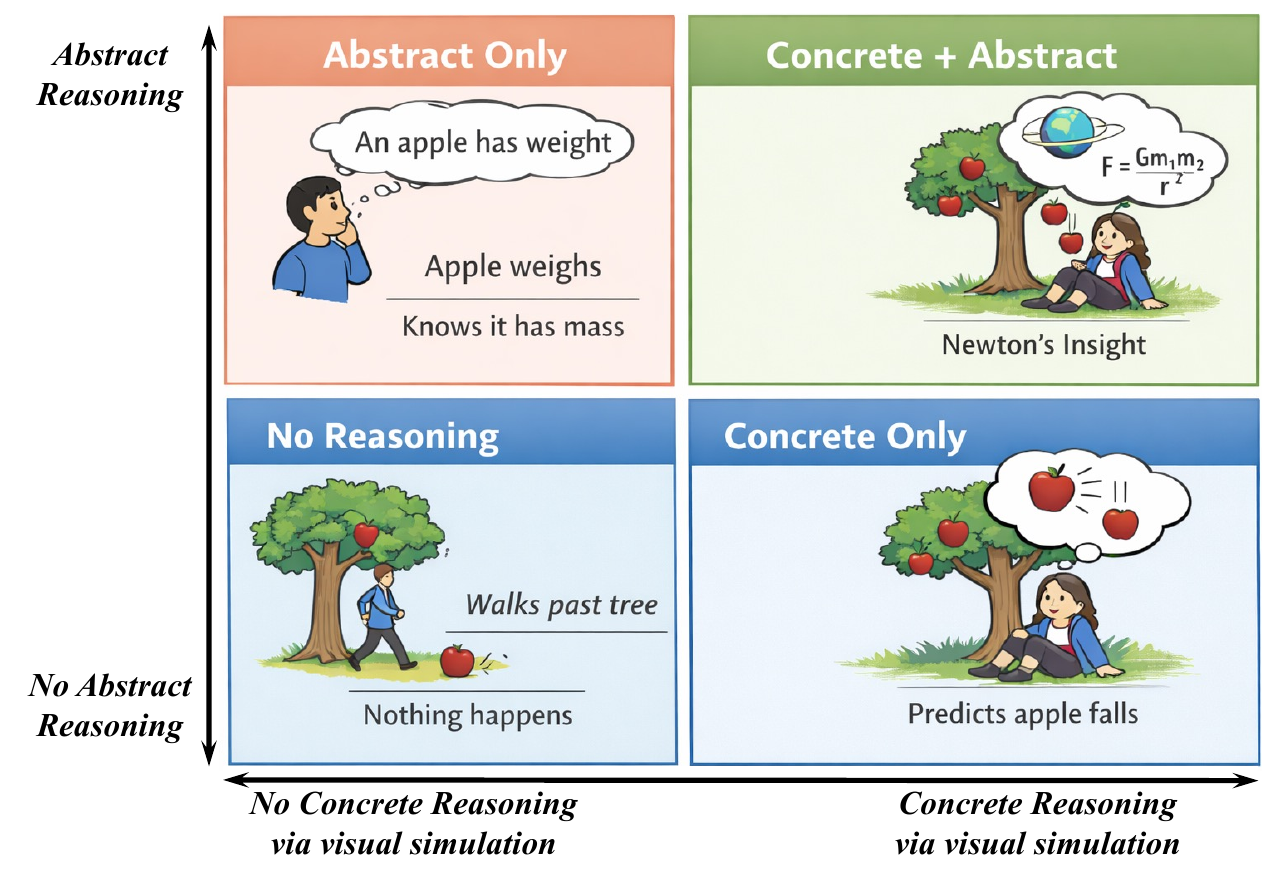}
\caption{\small Abstract reasoning organizes goals, rules, and questions in language, while concrete reasoning uses world-model rollouts to make possible futures visually explicit; reliable agents must coordinate both capabilities.}
\label{fig:intro}
\vspace{-15pt}
\end{wrapfigure}

Figure~\ref{fig:intro} shows this conceptual motivation: abstract language-level reasoning organizes goals, rules, and questions, while concrete rollout-based reasoning makes possible futures explicit. However, simply attaching a world model to an MLLM does not make this coordination reliable. Generated rollouts are noisy reasoning traces rather than precise oracles: they may miss task-critical interactions, drift from the initial geometry, or produce futures that are visually plausible but answer-incorrect \citep{qian2026current,Simulatingwm,Physicalwm}. The key challenge is therefore not whether world models can generate futures, but whether an MLLM can control when those futures should influence reasoning.

A preliminary empirical study makes this challenge concrete. Under optional tool use, models often continue to rely on abstract reasoning even when simulation would help, which we call \emph{Simulation Inertia}. Under forced simulation, models may accept misleading rollouts without sufficient scrutiny, leading to the \emph{Forced-Simulation Paradox}. These failures suggest that world-model assistance requires arbitration between abstract priors and rollout-based evidence, rather than unconditional generation or unconditional trust.

We formulate this problem as \emph{controlled concrete reasoning}: given an initial observation and a future-oriented question, the MLLM must learn when to invoke a world model, how to verify the resulting rollout, and how much to rely on it when predicting the final answer. The external world model supplies candidate visual futures, while the MLLM remains responsible for simulation selection, rollout verification, rollout reliance, and answer prediction.

To evaluate this setting, we construct two complementary human-verified benchmarks. \texttt{VRQABench} targets controllable spatial lookahead in maze, irregular-maze, and Sokoban-style puzzle environments \citep{VRBench}; \texttt{OpenWorldQA} targets open-domain physical prediction from pre-event anchor frames in real-world videos. Together, they test future prediction from static initial conditions across structured spatial environments and natural physical scenes.

We further propose \emph{Privileged-Future On-Policy Self-Distillation} (PF-OPSD), a training framework tailored to these simulation-control decisions. During training, a privileged evaluator observes the ground-truth future video and answer only as teacher-side context, scores the utility of the student's on-policy concrete-reasoning trajectories, and distills advantage-weighted targets back into a deployable student. At test time, the student has no access to true futures and must decide for itself when to simulate, verify, rely, or fall back to abstract reasoning.

Our contributions are summarized as follows:
\begin{itemize}[leftmargin=*, itemsep=1pt, topsep=1pt, partopsep=1pt, parsep=1pt]
    \item We frame future outcome prediction as controlled concrete reasoning, where an MLLM learns when to invoke world-model rollouts, how to verify them, and how much to rely on them alongside abstract reasoning.
    \item We identify \emph{Simulation Inertia} and \emph{Forced-Simulation Paradox} as two failure modes that reveal limits of naive world-model attachment.
    \item We construct two human-verified evaluation settings, \texttt{VRQABench} and \texttt{OpenWorldQA}, for controllable spatial lookahead and open-domain physical future prediction from initial observations.
    \item We propose PF-OPSD, which uses privileged future context during training to distill concrete-reasoning decisions into an MLLM, yielding gains of 10.5 and 10.8 percentage points over the baseline on two benchmarks.
\end{itemize}

\section{Related Work}

LLMs support structured inference and tool use through chain-of-thought, self-consistency, program-aided reasoning, and agentic tool invocation \citep{wei2022chain,wang2022self,KojimaGRMI22,chen2022program,yao2023react,schick2023toolformer}. MLLMs extend these abilities to image-text reasoning \citep{SurveyMLLM,VisualMLLM,qwen3vl}, while visual and physical reasoning benchmarks study compositional VQA, intuitive physics, and outcome prediction \citep{johnson2017clevr,hudson2019gqa,bakhtin2019phyre,riemer2020learning}.
Recent video models can serve as world models for future rollouts \citep{tong2022videomae,bardes2024vjepa,wiedemer2025video,wan2025,HunyuanVideo,Simulatingwm,Physicalwm}, but such rollouts may hallucinate or drift from task-relevant geometry \citep{luo2025vimo,cao2026mobiledreamer,qian2026current}. PF-OPSD treats these rollouts as noisy concrete-reasoning traces and instantiates distillation, privileged information, and on-policy learning \citep{hinton2015distilling,vapnik2009new,schulman2017proximal} for simulation-control decisions. The full related work is provided in Appendix~\ref{sec:appendix_related_work}.

\section{Problem Definition and Benchmarks}
\label{sec:problem_benchmarks}

\subsection{Controlled Concrete Reasoning}

We define world-model-assisted future prediction as a controlled concrete-reasoning problem. Given a current image or pre-event anchor frame $o$ and a question $q$, the agent predicts a future-outcome answer $y$ while optionally using a generative world model $W$. The world model provides concrete reasoning through candidate future rollouts: when invoked with an agent-written prompt, it returns a rollout $\hat{v}$ rather than an answer. The ground-truth future video $v^*$ is used only as privileged training information and is unavailable at test time.

The policy produces a concrete-reasoning trajectory instead of consuming $\hat{v}$ as a fixed input. It first decides whether simulation is needed,
\begin{align}
    d_{\mathrm{sim}} \sim \pi_\theta(\cdot \mid o, q),
\end{align}
and, if $d_{\mathrm{sim}}=1$, writes a simulation prompt and receives a rollout:
\begin{align}
    p_{\mathrm{sim}} \sim \pi_\theta(\cdot \mid o, q, d_{\mathrm{sim}}),  \hat{v} \sim W(\cdot \mid o, p_{\mathrm{sim}}).
\end{align}
The same policy then either verifies and relies on the rollout, or falls back to abstract reasoning when no simulation is used:
\begin{align}
    \!\!\!\!z_{\mathrm{ver}}, z_{\mathrm{rel}}, y
    &\!\sim\! \pi_\theta(\cdot \!\mid\! o, q, d_{\mathrm{sim}}, p_{\mathrm{sim}}, \hat{v}),\!\!\!\!
    && d_{\mathrm{sim}}\!=\!1 \nonumber \\
    \!\!\!\!z_{\mathrm{rel}}, y
    &\!\sim\! \pi_\theta(\cdot \!\mid\! o, q, d_{\mathrm{sim}}),\!\!\!\!
    && d_{\mathrm{sim}}\!=\!0\!\!
\end{align}
Here, $z_{\mathrm{ver}}$ is the rollout-verification decision and $z_{\mathrm{rel}}$ is the rollout-reliance or abstract-reasoning fallback process. The objective is to improve future prediction while avoiding negative transfer from erroneous simulations and unnecessary reliance on weak rollouts.

\subsection{Why Naive Integration Fails}

\begin{wrapfigure}{r}{0.5\linewidth}
\centering
\vspace{-10pt}
\includegraphics[width=\linewidth]{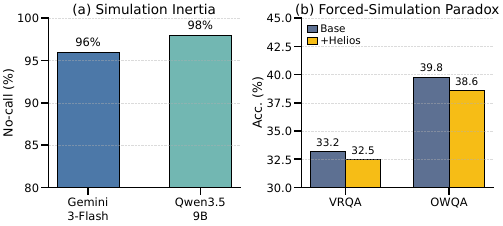}
\caption{\small Preliminary empirical observations showing simulation inertia and the forced-simulation paradox. Panel (a) reports the no-call probability under optional world-model use, and Panel (b) compares accuracy before and after forced Helios simulation.}
\label{fig:motivation}
\vspace{-10pt}
\end{wrapfigure}

We conduct a preliminary diagnostic on \texttt{VRQABench} using \texttt{Gemini-3-Flash} \citep{pichai2025new} as the language agent and \texttt{Helios} \citep{yuan2026helios} as the video world model. As shown in Figure~\ref{fig:motivation}, we compare optional world-model use with forced simulation to isolate two limitations of naive world-model attachment.
First, optional tool use leads to \emph{Simulation Inertia}: even when prompts encourage simulation for complex spatial reasoning, the agent often relies on abstract reasoning and does not call the world model. Second, forced simulation leads to a \emph{Forced-Simulation Paradox}: providing a generated rollout for every query does not necessarily improve accuracy and may even hurt performance when the rollout is visually plausible but task-incorrect.
These observations show that the core problem is not merely access to more future videos, but learning when to request, trust, discount, or reject noisy rollouts.

\subsection{Benchmark Suite for Future Prediction from Static Observations}

\begin{wraptable}{r}{0.52\linewidth}
\centering
\small
\setlength{\tabcolsep}{3pt}
\resizebox{\linewidth}{!}{%
\begin{tabular}{lllll}
\toprule
Benchmark & Type & Train & Test & Coverage \\
\midrule
\texttt{VRQABench} & Puzzles & 4,000 & 636 & 5 spatial categories \\
\texttt{OpenWorldQA} & Real-World & 3,904 & 500 & 12 physical categories \\
\bottomrule
\end{tabular}}
\caption{\small Overview of the two benchmark distributions. All examples are four-choice questions from an initial image or anchor frame; category definitions and split details are provided in Appendix~\ref{sec:dataset_details}, and construction prompts are provided in Appendix~\ref{sec:appendix}.}
\label{tab:dataset_overview}
\vspace{-6pt}
\end{wraptable}

We introduce two four-choice benchmarks for the same input regime: a model observes only an initial state and must answer a question about a later outcome. \texttt{VRQABench} isolates rule-governed spatial lookahead in puzzle environments, while \texttt{OpenWorldQA} tests open-domain physical prediction from natural videos. Future frames are never part of the benchmark input; generated rollouts, when used, are optional concrete-reasoning inputs controlled by the agent.

\paragraph{\texttt{VRQABench}: Controllable Spatial Lookahead from Initial Puzzle Images.}
\texttt{VRQABench} is built from VR-Bench \citep{VRBench} by turning maze, irregular-maze, and Sokoban states into multiple-choice future-prediction questions. For each puzzle, we first derive the target statistic from the underlying state with deterministic solvers: shortest-path search and geometric path analysis for maze variants, and Sokoban search for box-pushing tasks. We then use language models only for surface realization, writing the question, producing plausible distractors, and filtering item quality, so that labels remain programmatically grounded. After automatic filtering, human annotators verify every retained item for visual consistency, option plausibility, and answer validity; items that fail this final check are removed. The full construction prompt is provided in Appendix~\ref{sec:appendix}.

The resulting benchmark contains 4,636 human-verified questions, with 4,000 training examples and 636 evaluation examples. Its categories cover turn counting, turn direction, Sokoban pushes, direction counts, and push-direction counts; detailed category definitions and split distributions are provided in Appendix~\ref{sec:dataset_details}. For world-model-assisted experiments, \texttt{VRQABench} uses the VR-Bench-fine-tuned Helios world model \citep{yuan2026helios} only as an external rollout generator; the evaluated model still receives the initial image, question, options, and any invoked rollout.

\paragraph{\texttt{OpenWorldQA}: Predicting Real-World Physical Futures from Anchor Frames.}
\texttt{OpenWorldQA} uses short real-world videos but exposes only a pre-outcome anchor frame to the evaluated model. We construct it with a five-stage agentic pipeline. A scene-analysis stage selects anchor frames that contain enough initial-condition cues without revealing the outcome; a question-design stage writes one-to-three-step physical prediction questions; a distractor stage creates locally plausible alternatives; a small-model probe removes items that are too easy; and a reviewer verifies answer correctness, anchor validity, distractor plausibility, visual consistency, and category alignment. Each surviving item then undergoes human verification, where annotators check whether the anchor frame supports prediction, the future outcome is unambiguous, and the answer/options are physically valid; bad samples are filtered out. Detailed prompts for these stages are included in Appendix~\ref{sec:appendix}.

The resulting benchmark contains 4,404 human-verified four-choice questions, with 3,904 training examples and a 500-question balanced test set. It spans 12 physical-reasoning categories and six question forms, including order, count, first contact, intermediate state, failure, and counterfactual questions; Appendix~\ref{sec:dataset_details} gives the full category taxonomy and split distributions. Table~\ref{tab:dataset_overview} summarizes the benchmark distributions.

\section{Controlled Concrete Reasoning with PF-OPSD}
\label{sec:method}

\begin{figure*}[t]
\centering
\includegraphics[width=\textwidth]{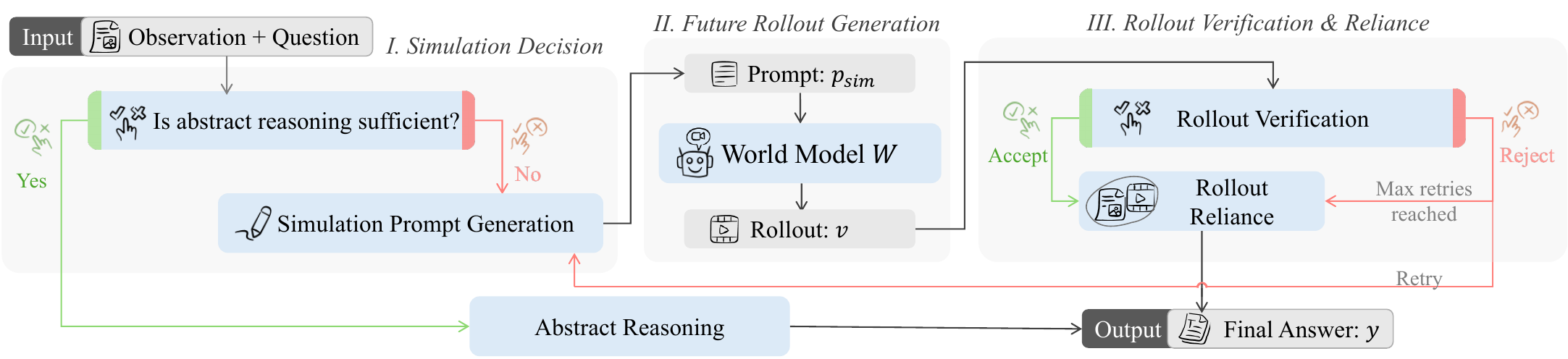}
\caption{\small Inference-time controlled concrete reasoning. The student MLLM first decides whether abstract reasoning is sufficient. If simulation is needed, it writes a simulation prompt $p_{\mathrm{sim}}$, queries the world model $W$ for a candidate rollout $\hat{v}$, verifies the rollout, and decides how much to rely on it alongside abstract reasoning. Rejected rollouts trigger prompt retry up to three simulation attempts; once this per-example cap is reached, the policy proceeds with its rollout-reliance state before predicting $y$.}
\label{fig:model}
\end{figure*}

Figure~\ref{fig:model} illustrates the overall PF-OPSD pipeline. The central difficulty is not access to the world model $W$, but controlling how uncertain rollouts are used for concrete reasoning. Given an input $x=(o,q,\mathcal{O})$, we define the induced trajectory distribution as follows:
\begin{align}
p_\theta(\tau\!\mid\! x;W)
\!\!=\!\!\!\!\!\! \prod_{t\in\mathcal{T}(\tau)}\!\!\!\!
\pi_\theta(a_t\mid h_t) \!\cdot \!\!\!\!\!
\prod_{i=1}^{N_{\mathrm{sim}}(\tau)}\!\!
\!\!W(\hat{v}^{(i)}\!\mid\! o,p_{\mathrm{sim}}^{(i)}).
\end{align}
Here, $\tau$ contains the control actions and final answer, $h_t$ is the history before action $a_t$, and $W$ only samples candidate futures. In our experiments, $W$ is instantiated by Helios \citep{yuan2026helios}: \texttt{VRQABench} uses the VR-Bench-fine-tuned Helios model, while \texttt{OpenWorldQA} uses the general Helios model.

PF-OPSD trains this deployable student policy with asymmetric information. During training, a privileged evaluator $E^+$ observes the ground-truth future $v^*$ and answer $y^*$ as teacher-side context, and uses them to score candidate concrete-reasoning trajectories generated by the student. During inference, $E^+$, $v^*$, and $y^*$ are removed; the student receives only $x$ and optional rollouts from $W$. The learning problem is
\begin{align}
J(\theta)
\!\!&=\mathbb{E}_{(x,y^*,v^*)}
\mathbb{E}_{\tau\sim p_\theta(\cdot\mid x;W)}
\!\left[R^+(\tau;y^*,v^*)\right], \nonumber\\[-1pt]
\theta^*
\!\!&=\arg\max_\theta J(\theta),
\quad (y^*,v^*)\notin x_{\mathrm{test}}.
\end{align}
Thus, privileged futures and answers are used only to construct training targets for the student's own trajectories, not as test-time inputs.

\subsection{Policy Trajectory and Action Space}

Given an observation $o$ and a question $q$, the student policy emits a concrete-reasoning trajectory rather than consuming a rollout as a fixed input. Without simulation, the trajectory is
\begin{align}
\tau_{\mathrm{no}}=(d_{\mathrm{sim}}, z_{\mathrm{rel}}, y).
\end{align}
When simulation is invoked, the trajectory contains a bounded sequence of simulation attempts:
\begin{align}
\!\!\!\tau_{\mathrm{sim}}
\!\!=\!\!(d_{\mathrm{sim}}, \!\mathcal{A}_{1:m}, \!z_{\mathrm{rel}}, y), \mathcal{A}_i
\!\!=\!\!(p_{\mathrm{sim}}^{(i)},\!\hat{v}^{(i)},\!z_{\mathrm{ver}}^{(i)})\!\!
\end{align}
where $1\leq m\leq B$ and $B=3$. Each attempt writes a prompt $p_{\mathrm{sim}}^{(i)}$, receives $\hat{v}^{(i)}\sim W(\cdot\mid o,p_{\mathrm{sim}}^{(i)})$, and predicts $z_{\mathrm{ver}}^{(i)}\in\{\text{accept},\text{reject},\text{uncertain}\}$. Let $a$ denote \text{accept}. The number of attempts is determined by
\begin{align}
\!\!m\!=\!\tau_{\mathrm{stop}}
\!\!=\!\!
\min\{i\leq B\!\!:\!\!\, z_{\mathrm{ver}}^{(i)}=a\ \vee\ i=B\}.\!\!
\end{align}
The full probability of this trajectory follows the opening product form, with one policy factor for each control action and one world-model factor for each queried rollout.

The action space exposes five control decisions:
\begin{itemize}[leftmargin=*, itemsep=1pt, topsep=1pt, partopsep=1pt, parsep=1pt]
    \item \textbf{Simulation decision.} $d_{\mathrm{sim}}\in\{\text{yes},\text{no}\}$ decides whether concrete reasoning from a rollout is likely to be useful.
    \item \textbf{Simulation query.} $p_{\mathrm{sim}}^{(i)}$ specifies the task-relevant objects, paths, contacts, or event changes for attempt $i$.
    \item \textbf{Rollout verification.} $z_{\mathrm{ver}}^{(i)}$ judges whether the rollout is consistent, plausible, and relevant to the question.
    \item \textbf{Rollout reliance.} $z_{\mathrm{rel}}$ states how an accepted, uncertain, or rejected rollout should be used, discounted, or overridden.
    \item \textbf{Answer prediction.} $y\in\{A,B,C,D\}$ is the final choice.
\end{itemize}
This trajectory makes the failure points explicit: the model can underuse useful simulation, accept misleading rollouts, or over-rely on weak rollouts.

\subsection{Two-Stage Training}

PF-OPSD separates format learning from utility calibration. Stage 1 teaches the student to produce valid concrete-reasoning trajectories. Stage 2 calibrates these trajectories with a teacher-side evaluator that sees the privileged future only during training.

\paragraph{Stage 1: protocol SFT.}
We initialize the student with protocol supervision from a \texttt{Gemini-3.1-Pro + Agent workflow} \citep{pichai2025new}. For each training example, the workflow observes the image or anchor frame, question, options, ground-truth answer, and training-time future video as teacher context, and generates a structured trajectory over $d_{\mathrm{sim}}$, $p_{\mathrm{sim}}$, $z_{\mathrm{ver}}$, $z_{\mathrm{rel}}$, and $y$. We keep only trajectories that pass rule checks, answer-consistency checks, and reviewer-style filtering. This stage fixes the output protocol; the workflow is not used during PF-OPSD self-distillation or test-time inference.

\paragraph{Stage 2: privileged-future on-policy self-distillation.}
The second stage calibrates decisions produced by the current student. For each training example, the student first generates a base trajectory $\tau_s$ under the student-view context $c=(o,q,\text{options})$, which excludes the ground-truth future $v^*$ and answer $y^*$. At each decision node $t$ with prefix $h_t^s$, we build a candidate set $C_t$. Discrete nodes use the valid action set, including $d_{\mathrm{sim}}$, $z_{\mathrm{ver}}^{(i)}$, and $y$. Text nodes, including $p_{\mathrm{sim}}^{(i)}$ and $z_{\mathrm{rel}}$, use $K$ samples from the current policy.

Each candidate action $a\in C_t$ is forced at node $t$, after which the remaining trajectory is greedily completed as $\tau_a$. The privileged evaluator $E^+$ then receives $(c_t,h_t^s,a,\tau_a,y^*,v^*)$ and assigns a teacher-side score. In our implementation, $E^+$ is instantiated by \texttt{Qwen3.6-27B}, a native multimodal model that observes the generated rollout and the ground-truth future video only during training. Concretely, $E^+$ uses $y^*$ to judge final-answer correctness and uses $v^*$ to judge whether accepted rollouts are consistent with the actual future and useful for the question:
\begin{align}
\!\!R^+(\tau_a)
&={} \mathbf{1}[\hat{y}=y^*]
- \lambda_{\mathrm{sim}}N_{\mathrm{sim}}  \\
& - \lambda_{\mathrm{FA}}\mathbf{1}[\text{false accept}] \!-\! \lambda_{\mathrm{FR}}\mathbf{1}[\text{false reject}].\!\!\nonumber
\end{align}
These labels are evaluator-derived training signals, not human oracle annotations; their operational definitions are provided in Appendix Table~\ref{tab:evaluator_labels}. The evaluator also provides a teacher-view preference $P_t^+(a)=\pi^+(a\mid c_t^+,h_t^s)$ under $c_t^+=c_t\cup\{y^*,v^*\}$. We set $Q_t^+(a)=R^+(\tau_a)$ and compute $A_t^+(a)=Q_t^+(a)-V_t^+$, where
\begin{align}
V_t^+
\!\!&=
\begin{cases}
\sum_{a\in C_t}\!P_t^+(a)Q_t^+(a), & t\in\mathcal{C}_{\mathrm{disc}}, \\
K^{-1}\sum_{k=1}^{K}Q_t^+(a_k), & t\in\mathcal{C}_{\mathrm{text}}.
\end{cases}
\end{align}
This on-policy design asks which alternatives would improve the student's own behavior, rather than imitating a fixed privileged trace. Appendix~\ref{sec:pfopsd_algorithm} gives the procedural form.

\subsection{Advantage-Weighted Distillation Objective}

The privileged advantages define student-view targets. For a discrete node $t$, we form
\begin{align}
q_t^\star(a)
\!\!&=\!Z_t^{-1}\pi^+(a\mid c_t^+,h_t^s)
\exp(A_t^+(a)/\tau_A), \!\!\\
Z_t
\!\!&=\!\sum_{a'\in C_t}\!\pi^+(a'\mid c_t^+,h_t^s)
\exp(A_t^+(a')/\tau_A). \nonumber
\end{align}
The student minimizes
\begin{align}
\!\!\mathcal{L}_{\mathrm{disc}}
\!\!=\!\!\!\! \sum_{t\in\mathcal{C}_{\mathrm{disc}}}\!\!\!\!
D_{\mathrm{KL}}\Big(
\operatorname{sg}[q_t^\star(\cdot)] \!\,\|\,
\pi_\theta(\cdot\mid c_t,h_t^s)
\Big),\!\!
\end{align}
where $c_t$ excludes $y^*$ and $v^*$. For text nodes, we convert the privileged advantages into normalized candidate weights and optimize a weighted log-likelihood:
\begin{align}
w_{t,k}
\!\!&=\frac{\exp(A_t^+(a_k)/\tau_A)}
{\sum_j\!\exp(A_t^+(a_j)/\tau_A)}, \\
\!\!\mathcal{L}_{\mathrm{text}}
\!\!&=\! -\!\!\!\!\sum_{t\in\mathcal{C}_{\mathrm{text}}}\!\!\!\sum_{k=1}^{K}
\!\operatorname{sg}[w_{t,k}]\log\pi_\theta(a_k\!\mid\! c_t,h_t^s).\!\!
\end{align}
Combining the discrete-node KL objective and the text-node weighted likelihood gives the advantage-distillation loss:
\begin{align}
\mathcal{L}_{\mathrm{PF\text{-}OPSD}}^{\mathrm{adv}}
=
\mathcal{L}_{\mathrm{disc}}+
\mathcal{L}_{\mathrm{text}}.
\end{align}
{
The differentiable training objective combines protocol SFT with advantage-weighted distillation:
\begin{align}
\mathcal{L}
= \mathcal{L}_{\mathrm{SFT}}
+ \mathcal{L}_{\mathrm{PF\text{-}OPSD}}^{\mathrm{adv}}.
\end{align}
The discrete simulation cost is not added as a pathwise-differentiable count penalty. Instead, $-\lambda_{\mathrm{sim}}N_{\mathrm{sim}}$ in $R^+$ shapes the advantages and thereby the KL targets and text-action weights.
}
Unlike outcome-level RL such as GRPO \citep{shao2024deepseekmath}, PF-OPSD assigns credit to intermediate concrete-reasoning decisions, such as whether to simulate, whether to reject a hallucinated rollout, and whether to rely on abstract reasoning.

\subsection{Inference}

At test time, both $v^*$ and the protocol-generation workflow are removed. Deployment follows a learned simulation-control policy with only the student-view state $s_i=(x,\mathcal{A}_{1:i-1})$ and a hard per-example retry cap:
\begin{align}
\hat{\tau}
\!\!&=\arg\!\!\!\max_{\tau:\,N_{\mathrm{sim}}(\tau)\leq B}\!\!\!
\log p_\theta(\tau\mid x;W), B=3.\!\!
\end{align}
The first action gates simulation:
\begin{align}
d_{\mathrm{sim}}
\!\!&=\arg\max_{d\in\{\text{yes},\text{no}\}}
\pi_\theta(d\mid x).
\end{align}
If $d_{\mathrm{sim}}=\text{no}$, the policy answers from abstract reasoning. Otherwise, each attempt samples or decodes $p_{\mathrm{sim}}^{(i)}$, queries $W$ for $\hat{v}^{(i)}$, and predicts $z_{\mathrm{ver}}^{(i)}$. The retry gate is
\begin{align}
g_i
\!\!&=\mathbf{1}[z_{\mathrm{ver}}^{(i)}\neq\text{accept}]\,\mathbf{1}[i<B],
\end{align}
so rejected or uncertain rollouts trigger another prompt only before the per-example cap is reached. After stopping at $m$, the model forms a reliance state
\begin{align}
\!\!\!r_m
\!\!=\!\phi_\theta(x,\mathcal{A}_{1:m},z_{\mathrm{rel}}), \hat{y}
\!\!=\!\arg\max_y \pi_\theta(y\!\mid\! r_m)\!\!\!
\end{align}
This formulation makes inference a closed-loop policy over simulation selection, rollout verification, retry, rollout reliance, and answering. The world model remains a fallible concrete-reasoning source: rejected rollouts are not discarded mechanically, but can be discounted in $z_{\mathrm{rel}}$ when no accepted rollout is available.

\section{Experiments}

\subsection{Experimental Setup}

We evaluate PF-OPSD on the two future-prediction benchmarks introduced in Section~\ref{sec:problem_benchmarks}: \texttt{VRQABench} and \texttt{OpenWorldQA}. Unless otherwise stated, the student is \texttt{Qwen3.5-9B}~\citep{qwen3.5}; the external world model is \texttt{Helios}~\citep{yuan2026helios}, using the VR-Bench-fine-tuned Helios model for \texttt{VRQABench} and the general Helios model for \texttt{OpenWorldQA}; and protocol trajectories are generated offline by a privileged \texttt{Gemini-3.1-Pro + Agent} teacher~\citep{pichai2025new}. We train with Stage-1 protocol SFT followed by Stage-2 PF-OPSD self-distillation, and evaluate against zero-shot/no-simulation MLLMs, Qwen-based training baselines, and our workflow-agent prompting baseline using accuracy, simulation-decision quality, and rollout-verification metrics. Detailed splits, backbone/world-model choices, optimization hyperparameters, inference protocol, baseline definitions, and metric implementations are provided in Appendix~\ref{sec:experimental_setup_details}; dataset category taxonomies and distributions are in Appendix~\ref{sec:dataset_details}.

\subsection{Main Results}

Table~\ref{tab:vrqa} summarizes the benchmark results. PF-OPSD achieves the best accuracy on both benchmarks (72.3\% on \texttt{VRQABench} and 70.4\% on \texttt{OpenWorldQA}), improving over SFT and SFT + GRPO. The SFT baseline is already a learned-controller baseline because it is trained on structured trajectories containing simulation decisions, simulation prompts, rollout verification, rollout reliance, and final answers. The prompted \texttt{Workflow Agent} baseline, which has prompt-level access to Helios but no PF-OPSD training, performs worse than the image-only supervised model, indicating that world-model access alone is insufficient without learned simulation selection and rollout verification. The gains are consistent across the two benchmark designs rather than being driven by a single dataset: relative to SFT, PF-OPSD improves by 10.5 percentage points on \texttt{VRQABench} and 10.8 percentage points on \texttt{OpenWorldQA}. This suggests that the learned controller is not merely increasing world-model usage, but learning when rollouts are likely to be useful and when they should be rejected or discounted.

\begin{table}[t]
\centering
\small
\begin{tabular}{lcc}
\toprule
Model / Method & \texttt{VRQABench} & \texttt{OpenWorldQA} \\
\midrule
\multicolumn{3}{l}{\emph{Zero-shot / no-simulation baselines}} \\
\texttt{Gemini-3-Flash}~\citep{pichai2025new} & 45.9\% & 48.2\% \\
\texttt{GPT-5.4}~\citep{openai2026gpt54} & 43.2\% & 53.4\% \\
\texttt{HY3}~\citep{tencent2026hy3preview} & 38.2\% & 35.0\% \\
\texttt{Qwen3.6-27B}~\citep{qwen3.6-27b} & 33.0\% & 41.4\% \\
\texttt{Qwen3.5-9B}~\citep{qwen3.5} & 33.2\% & 39.8\% \\
\texttt{Qwen2.5-VL-7B}~\citep{qwen3vl} & 32.7\% & 14.2\% \\
\midrule
\multicolumn{3}{l}{\emph{Workflow-agent baselines}} \\
\texttt{Workflow Agent (Qwen3.5-9B)} & 32.5\% & 38.6\% \\
\texttt{Workflow Agent (Gemini-3-Flash)} & 47.2\% & 49.6\% \\
\midrule
\multicolumn{3}{l}{\emph{Qwen3.5-9B training baselines}} \\
\texttt{SFT}~\citep{qwen3.5} & 61.8\% & 59.6\% \\
\texttt{GRPO}~\citep{shao2024deepseekmath} & 63.5\% & 61.2\% \\
\texttt{PF-OPSD} & 72.3\% & 70.4\% \\
\bottomrule
\end{tabular}%
\caption{\small Overall accuracy on the two benchmarks. Full per-category results are reported in Table~\ref{tab:vrqa_detail} and Table~\ref{tab:openworld_detail} in the appendix.}
\label{tab:vrqa}
\end{table}
\begin{table}[t]
\centering
\small
\begin{tabular}{lllll}
\toprule
Variant & VRQA & OWQA & Call Rate & Calls \\
\midrule
\texttt{PF-OPSD} & 72.3\% & 70.4\% & 42.5\% & 0.45 \\
\midrule
w/o Simulation decision & 68.6\% & 67.2\% & 100\% & 1.00 \\
w/o Rollout verification & 65.3\% & 64.8\% & 42.5\% & 0.45 \\
w/o Rollout reliance & 67.8\% & 66.4\% & 42.5\% & 0.45 \\
w/o Advantage weighting & 66.4\% & 65.2\% & 45.2\% & 0.48 \\
answer-only distillation & 64.5\% & 63.8\% & 35.4\% & 0.38 \\
SFT & 61.8\% & 59.6\% & 23.0\% & 0.23 \\
\bottomrule
\end{tabular}
\caption{\small Ablation results for training mechanisms and concrete-reasoning nodes. Call Rate denotes the percentage of examples that trigger world-model simulation, and Calls denotes the average number of simulation calls per example.}
\label{tab:ablation}
\end{table}

\subsection{Ablation Studies}

Table~\ref{tab:ablation} isolates the contribution of each PF-OPSD component. SFT removes Stage-2 self-distillation but keeps the same structured reasoning-chain supervision, so it serves as a learned-controller cold start. Forcing simulation for all samples exposes the model to unnecessary noisy rollouts. Removing rollout verification or reliance weakens the model's ability to assess and balance concrete evidence, and removing advantage weighting reduces training to uncalibrated teacher-view imitation. Answer-only distillation further shows that supervising the final answer alone is insufficient.
Overall, the gains come from utility-calibrated supervision over intermediate concrete-reasoning actions, especially rollout verification and advantage weighting, rather than from simply adding generated videos.

\begin{figure}[t]
\centering
\begin{minipage}[t]{0.48\linewidth}
\centering
\includegraphics[width=\linewidth]{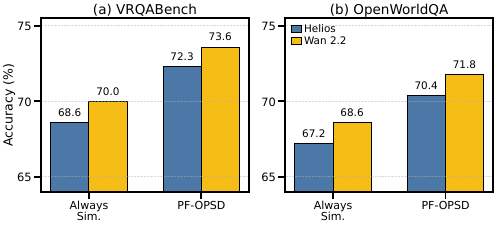}
\captionof{figure}{\small Cross-world-model diagnostic. Replacing \texttt{Helios} with \texttt{Wan 2.2} improves both the always-simulate variant and PF-OPSD, while PF-OPSD remains stronger than always simulating with the stronger rollout generator.}
\label{fig:cross_world_model}
\end{minipage}\hfill
\begin{minipage}[t]{0.48\linewidth}
\centering
\includegraphics[width=\linewidth]{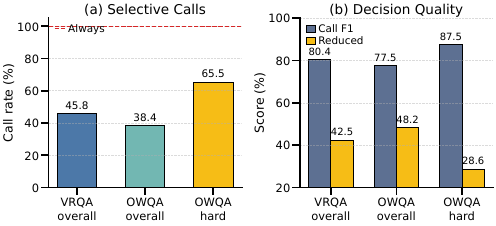}
\captionof{figure}{\small Simulation decision quality. PF-OPSD keeps the overall call rate well below always-simulate while increasing calls on OWQA hard samples and maintaining high agreement with privileged evaluator-derived simulation-help labels. ``Reduced'' denotes the fraction of unnecessary simulation calls avoided relative to an always-simulate policy, not the complement of Call Rate.}
\label{fig:decision}
\end{minipage}
\end{figure}

\subsection{Cross-World-Model Diagnostic}

We test cross-world-model transfer by replacing the default \texttt{Helios} generator with \texttt{Wan 2.2} at inference time while keeping the student policy and prompts unchanged. Figure~\ref{fig:cross_world_model} compares PF-OPSD with an \emph{Always Simulate (No Gate)} variant, which preserves the trained policy but forces one world-model query for every example. \texttt{Wan 2.2} gives slightly higher accuracy but is substantially slower, so we use it only as a diagnostic rollout generator. Its gains confirm that stronger world models can provide better concrete evidence, but the always-simulate variant with \texttt{Wan 2.2} still trails PF-OPSD with \texttt{Helios}, showing that world-model quality does not replace learned simulation control.

\subsection{Simulation Decision Analysis}

\begin{figure}[t]
\centering
\begin{minipage}[t]{0.48\linewidth}
\centering
\includegraphics[width=\linewidth]{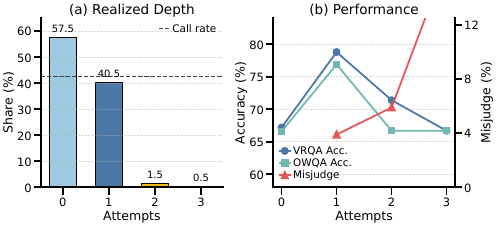}
\captionof{figure}{\small Realized simulation depth and performance under PF-OPSD. Attempts denotes the actual number of world-model calls selected by the policy for an example, not an externally imposed global constraint. Rows with one to three attempts in panel (a) sum to the learned call rate; panel (b) shows that repeated retries correspond to harder examples with lower accuracy and higher misjudgment.}
\label{fig:call_depth}
\end{minipage}\hfill
\begin{minipage}[t]{0.5\linewidth}
\centering
\includegraphics[width=\linewidth]{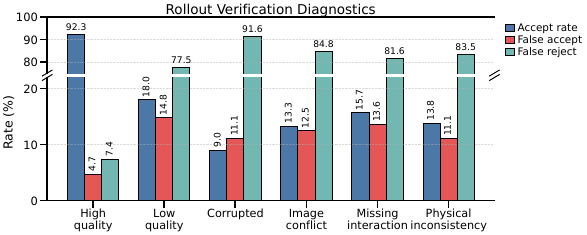}
\captionof{figure}{\small Rollout-verification behavior under controlled rollout-quality conditions. Accept rate measures how often PF-OPSD trusts a generated rollout, while false accept and false reject measure erroneous acceptance of misleading rollouts and erroneous rejection of useful rollouts, respectively. The broken y-axis is used only to improve the visibility of low-frequency errors, with exact values annotated on each bar.}
\label{fig:acceptance}
\end{minipage}
\end{figure}

\begin{figure}[t]
\centering
\begin{minipage}[t]{0.48\linewidth}
\centering
\includegraphics[width=\linewidth]{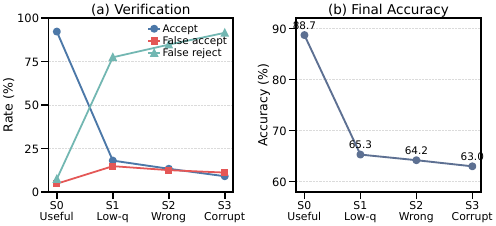}
\captionof{figure}{\small Rollout-quality severity diagnostic. PF-OPSD sharply reduces rollout acceptance as generated futures become less useful or more conflicting, while maintaining non-collapsed final accuracy under misleading rollouts.}
\label{fig:severity}
\end{minipage}\hfill
\begin{minipage}[t]{0.48\linewidth}
\centering
\includegraphics[width=\linewidth]{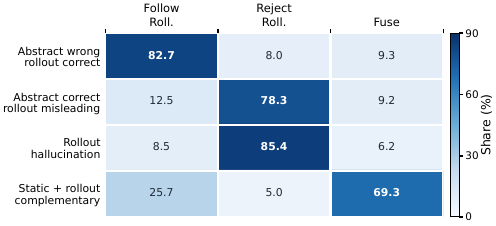}
\captionof{figure}{\small Representative conflict-resolution behavior. PF-OPSD follows a rollout when it corrects abstract reasoning, rejects misleading or hallucinated rollouts, and combines rollouts with static cues when the two sources are complementary.}
\label{fig:conflict_analysis}
\end{minipage}
\end{figure}

Figure~\ref{fig:decision} evaluates simulation selection using call rate, agreement with privileged evaluator-derived ``simulation-helps'' labels, and avoided calls relative to always simulating; detailed subset results are provided in Appendix Table~\ref{tab:decision}. PF-OPSD invokes the world model selectively, calling more often for dynamically or spatially demanding samples and avoiding calls when static cues are sufficient.
Figure~\ref{fig:call_depth} further breaks down the realized number of simulation attempts. The aggregate behavior matches Table~\ref{tab:ablation}: 42.5\% of examples invoke simulation, with 0.45 calls per example on average. Multi-attempt cases correspond to harder examples where earlier rollouts are rejected or uncertain, resulting in lower accuracy and higher misjudgment.

\subsection{Rollout Verification Analysis}

Figure~\ref{fig:acceptance} summarizes whether PF-OPSD decides to trust generated rollouts under controlled rollout-quality conditions. The model accepts high-quality rollouts at a high rate, while sharply reducing acceptance for corrupted, conflicting, incomplete, or physically inconsistent rollouts. The full diagnostic table, including verification precision, recall, false acceptance, false rejection, and final accuracy, is provided in Appendix Table~\ref{tab:acceptance}.
Figure~\ref{fig:severity} compresses these controlled conditions into a rollout-quality severity diagnostic. As rollout quality degrades from verified useful futures to incomplete, wrong-but-plausible, and strongly corrupted rollouts, PF-OPSD reduces its acceptance rate from 92.3\% to 9.0\%. Final accuracy degrades gracefully rather than collapsing, suggesting that the policy can discount unreliable concrete evidence and fall back on abstract reasoning.

\subsection{Rollout Reliance and Conflict Analysis}

Figure~\ref{fig:conflict_analysis} summarizes representative diagnostic subsets in which abstract reasoning and rollouts are conflicting or complementary. It visualizes whether the model follows the rollout, rejects it, or combines it with static cues; the full subset-level table, including final accuracy and recovery, is provided in Appendix Table~\ref{tab:conflict}.
The goal of PF-OPSD is not to always trust simulation. Rather, the model should use rollouts when they provide corrective concrete reasoning, reject them when they are misleading, and combine them with static cues when the two sources are complementary. The results show that the model recovers many errors when abstract reasoning is wrong but the rollout is correct, and that it often rejects hallucinated or misleading rollouts when static cues are stronger.

\section{Conclusion}

We studied multimodal future outcome prediction from current images or pre-event anchor frames and framed world-model assistance as controlled concrete reasoning rather than naive world-model attachment. To evaluate this setting, we introduced human-verified \texttt{VRQABench} and \texttt{OpenWorldQA}, covering controllable spatial planning and open-domain physical prediction. We further proposed PF-OPSD, which uses training-time future videos and answers as privileged teacher-side context to calibrate simulation, rollout verification, rollout reliance, and answer decisions. Experiments show improved accuracy and stronger robustness to noisy or conflicting rollouts.

\section*{Limitations}

This work focuses on image-conditioned future prediction with optional concrete reasoning from generative world models. Its conclusions are most directly applicable to settings where the world model can produce futures that are at least partially relevant to the given scene and question. When generated rollouts are poorly aligned with the input, PF-OPSD is intended to reduce unnecessary reliance on them rather than to replace improvements in world-model quality.

Our experiments cover two complementary benchmarks and a concrete world-model interface, but they do not exhaust all forms of physical reasoning or all possible world-model designs. More specialized domains, longer temporal horizons, and interactive environments may require additional benchmark coverage and task-specific prompting strategies.

Finally, PF-OPSD uses privileged future videos only during training to estimate action utility. This keeps the deployed policy free from ground-truth future inputs, while making the approach depend on the availability and alignment of training-time privileged signals. Extending the framework to weaker or more implicit forms of future supervision is a promising direction for future work.

\bibliography{custom}

\clearpage
\appendix

\section{Detailed Experimental Setup}
\label{sec:experimental_setup_details}

\paragraph{Datasets and splits.}
All training and evaluation use the splits introduced in Section~\ref{sec:problem_benchmarks}: \texttt{VRQABench} provides 4{,}000 training and 636 evaluation puzzles over 5 spatial categories, and \texttt{OpenWorldQA} provides 3{,}904 training and 500 evaluation anchor-frame questions over 12 physical categories. Both datasets are filtered by automatic reviewers and then manually verified item by item; retained examples must have a valid initial observation, a unique answer, plausible distractors, and no visible leakage of the future outcome. Stage-1 protocol SFT and Stage-2 PF-OPSD self-distillation are run on the union of the two training splits; all reported numbers are computed on the held-out evaluation splits, which are never seen during training. Detailed category taxonomies and per-split distributions are given in Appendix~\ref{sec:dataset_details}.

\paragraph{Backbone, world model, and protocol teacher.}
Unless otherwise stated, the student MLLM is \texttt{Qwen3.5-9B}~\citep{qwen3.5}; the external world model is \texttt{Helios}~\citep{yuan2026helios}. We use the VR-Bench-fine-tuned Helios model for \texttt{VRQABench}, and the general Helios model for \texttt{OpenWorldQA}; in both cases, Helios is queried only when the policy emits $d_{\mathrm{sim}}=1$. Stage-1 protocol trajectories are produced by a \texttt{Gemini-3.1-Pro + Agent} workflow~\citep{pichai2025new} that sees the privileged future video as teacher-side context; this workflow is used only for offline data generation and is not invoked during PF-OPSD training or at test time.

\paragraph{Training.}
Following the compact post-training configuration reported in OPSD~\citep{zhao2026selfdistilledreasoner}, we tune only LoRA adapters on the student MLLM, with rank $r=64$, scaling factor $\alpha=128$, and target modules \texttt{q\_proj}, \texttt{k\_proj}, \texttt{v\_proj}, \texttt{o\_proj}, \texttt{gate\_proj}, \texttt{up\_proj}, and \texttt{down\_proj}; the visual encoder and external world model are frozen. Both stages use AdamW with learning rate $5\times10^{-6}$, effective batch size 32, bfloat16 precision, gradient checkpointing, and FlashAttention-2. Stage-1 SFT optimizes $\mathcal{L}_{\mathrm{SFT}}$ on filtered protocol trajectories for one epoch, with maximum sequence length 16k. {Stage-2 PF-OPSD runs for 100 on-policy update steps and optimizes $\mathcal{L}=\mathcal{L}_{\mathrm{SFT}}+\mathcal{L}^{\mathrm{adv}}_{\mathrm{PF\text{-}OPSD}}$. The simulation cost is incorporated through $-\lambda_{\mathrm{sim}}N_{\mathrm{sim}}$ in the privileged reward rather than through a separate differentiable loss. Unless otherwise stated, we use $(\lambda_{\mathrm{sim}},\lambda_{\mathrm{FA}},\lambda_{\mathrm{FR}})=(0.05,0.50,0.25)$, advantage temperature $\tau_A=0.5$, and $K=4$ candidate samples at text nodes;} text candidates are sampled with temperature 1.1 and forced rollouts are greedily completed. The same optimizer and adapter settings are used for all trainable Qwen-based baselines, changing only the training objective.

\paragraph{Inference.}
At test time the student receives only the image (or anchor frame), question, and options; the privileged future $v^*$, answer $y^*$, and protocol teacher are removed. The policy decodes a structured trajectory over $(d_{\mathrm{sim}},p_{\mathrm{sim}},z_{\mathrm{ver}},z_{\mathrm{rel}},y)$ with a hard per-example simulation cap $B=3$, following the inference objective in Section~\ref{sec:method}. Decoding is deterministic (greedy) for discrete control nodes and uses a low temperature for text nodes.

\paragraph{Baselines.}
We compare against three families covering the natural alternatives to controlled integration: (i) \textbf{Zero-shot / no-simulation MLLMs}, including closed-source reference models such as \texttt{Gemini-3-Flash}, \texttt{GPT-5.4}, and \texttt{HY3}, as well as Qwen-series MLLMs that answer directly from the image and question; (ii) \textbf{Qwen-based training baselines}, including \texttt{SFT} and \texttt{SFT+GRPO}~\citep{shao2024deepseekmath} on the same backbone and data; and (iii) \textbf{Workflow-agent baselines}, where either \texttt{Qwen3.5-9B} or \texttt{Gemini-3-Flash} is given prompt-level access to Helios rollouts but receives no PF-OPSD training. SFT is trained on the same structured trajectory format as PF-OPSD, but without on-policy privileged utility calibration.

\paragraph{Metrics.}
We report (1) overall and per-category multiple-choice accuracy on both benchmarks; (2) average simulation calls per sample $N_{\mathrm{sim}}$ and call rate; (3) decision quality via call precision/recall/F1 against privileged evaluator-derived ``simulation-helps'' labels; and (4) rollout verification precision/recall together with false-accept and false-reject rates under controlled rollout-quality conditions. Unless noted, all numbers are computed on the fixed evaluation splits.

\paragraph{Privileged evaluator-derived labels.}
Table~\ref{tab:evaluator_labels} summarizes the operational labels produced by the privileged evaluator. These labels are derived by comparing the generated rollout with the ground-truth future using \texttt{Qwen3.6-27B}; they are not human oracle annotations.

\begin{table}[ht]
\centering
\small
\setlength{\tabcolsep}{3pt}
\begin{tabular}{p{0.24\linewidth}p{0.68\linewidth}}
\toprule
Label & Operational definition \\
\midrule
Useful rollout & The evaluator judges that the generated rollout is consistent with $v^*$ on the task-relevant future outcome and informative for answering $q$. \\
False accept & The student accepts a rollout that the evaluator marks as inconsistent with $v^*$ or not useful for answering $q$. \\
False reject & The student rejects a rollout that the evaluator marks as consistent with $v^*$ and useful for answering $q$. \\
Simulation helps & A diagnostic label indicating that rollout evidence can correct or materially support the final answer. \\
\bottomrule
\end{tabular}
\caption{\small Operational definitions of privileged evaluator-derived labels.}
\label{tab:evaluator_labels}
\end{table}

\begin{table*}[ht]
\centering
\small
\setlength{\tabcolsep}{4pt}
\begin{tabular}{lllllllll}
\toprule
Dataset / Subset & \# & Call Rate & Precision & Recall & F1 & Acc. Called & Acc. Not Called & Reduced \\
\midrule
\texttt{VRQABench} overall & 636 & 45.8\% & 82.5\% & 78.4\% & 80.4\% & 78.4\% & 67.2\% & 42.5\% \\
\texttt{VRQABench} maze & 275 & 38.9\% & 85.0\% & 82.0\% & 83.5\% & 74.8\% & 66.7\% & 52.4\% \\
\texttt{VRQABench} irregular & 171 & 45.6\% & 87.2\% & 86.1\% & 86.6\% & 74.4\% & 63.4\% & 48.5\% \\
\texttt{VRQABench} sokoban & 190 & 55.8\% & 76.4\% & 69.8\% & 73.0\% & 84.9\% & 72.6\% & 35.3\% \\
\texttt{OpenWorldQA} overall & 500 & 38.4\% & 79.7\% & 75.4\% & 77.5\% & 76.6\% & 66.6\% & 48.2\% \\
OWQA hard & 84 & 65.5\% & 89.1\% & 86.0\% & 87.5\% & 72.7\% & 55.2\% & 28.6\% \\
\bottomrule
\end{tabular}
\caption{\small Detailed simulation decision quality. The model preferentially invokes simulation for dynamically or spatially demanding samples.}
\label{tab:decision}
\end{table*}

\begin{table*}[ht]
\centering
\small
\begin{tabular}{llllllll}
\toprule
Rollout Condition & \# & Accept Rate & Precision & Recall & False Accept & False Reject & Acc. \\
\midrule
High-quality rollout & 300 & 92.3\% & 95.3\% & 92.6\% & 4.7\% & 7.4\% & 88.7\% \\
Low-quality rollout & 150 & 18.0\% & 85.2\% & 22.5\% & 14.8\% & 77.5\% & 65.3\% \\
Corrupted rollout & 100 & 9.0\% & 88.9\% & 8.4\% & 11.1\% & 91.6\% & 63.0\% \\
Conflicts with image cues & 120 & 13.3\% & 87.5\% & 15.2\% & 12.5\% & 84.8\% & 64.2\% \\
Misses key interaction & 140 & 15.7\% & 86.4\% & 18.4\% & 13.6\% & 81.6\% & 63.6\% \\
Physical inconsistency & 130 & 13.8\% & 88.9\% & 16.5\% & 11.1\% & 83.5\% & 64.6\% \\
\bottomrule
\end{tabular}
\caption{\small Full rollout-verification diagnostics under controlled rollout-quality conditions. Accept Rate measures how often PF-OPSD trusts the generated rollout; Precision and Recall evaluate verification decisions against privileged evaluator-derived rollout usefulness labels. False Accept is one minus precision, and False Reject is one minus recall. Final accuracy is task-answer accuracy and is computed independently of the verification confusion matrix.}
\label{tab:acceptance}
\end{table*}

\begin{table*}[ht]
\centering
\small
\begin{tabular}{lllllll}
\toprule
Conflict Subset & \# & Follow Roll. & Reject Roll. & Fuse & Final Acc. & Recovery \\
\midrule
Abstract wrong, rollout correct & 150 & 82.7\% & 8.0\% & 9.3\% & 85.3\% & 78.7\% \\
Abstract correct, rollout misleading & 120 & 12.5\% & 78.3\% & 9.2\% & 81.7\% & 75.0\% \\
Both uncertain & 100 & 45.0\% & 16.0\% & 39.0\% & 62.0\% & 56.0\% \\
Rollout hallucination & 130 & 8.5\% & 85.4\% & 6.2\% & 76.9\% & 72.3\% \\
Static cues insufficient & 160 & 75.6\% & 12.5\% & 11.9\% & 78.8\% & 74.4\% \\
Static cues and rollout complementary & 140 & 25.7\% & 5.0\% & 69.3\% & 87.9\% & 85.7\% \\
\bottomrule
\end{tabular}
\caption{\small Full conflict analysis between abstract reasoning and rollouts. Follow Roll., Reject Roll., and Fuse measure how PF-OPSD arbitrates between abstract reasoning and rollout-based concrete reasoning; Final Acc. and Recovery report task accuracy and recovered cases as percentages of each diagnostic subset.}
\label{tab:conflict}
\end{table*}

\begin{table*}[ht]
\centering
\small
\begin{tabular}{lllllll}
\toprule
Model / Method & Overall & C1 & C2 & C3 & C4 & C5 \\
\midrule
\multicolumn{7}{l}{\emph{Zero-shot / no-simulation baselines}} \\
\texttt{Gemini-3-Flash}~\citep{pichai2025new} & 45.9\% & 43.5\% & 41.2\% & 56.5\% & 46.3\% & 37.2\% \\
\texttt{GPT-5.4}~\citep{openai2026gpt54} & 43.2\% & 48.4\% & 46.1\% & 38.8\% & 31.6\% & 51.2\% \\
\texttt{HY3}~\citep{tencent2026hy3preview} & 38.2\% & 38.7\% & 47.9\% & 29.3\% & 30.5\% & 46.5\% \\
\texttt{Qwen3.6-27B}~\citep{qwen3.6-27b} & 33.0\% & 34.4\% & 31.5\% & 36.7\% & 27.4\% & 32.6\% \\
\texttt{Qwen3.5-9B}~\citep{qwen3.5} & 33.2\% & 41.4\% & 44.2\% & 25.9\% & 11.6\% & 27.9\% \\
\texttt{Qwen2.5-VL-7B}~\citep{qwen3vl} & 32.7\% & 29.0\% & 44.8\% & 21.1\% & 34.7\% & 37.2\% \\
\midrule
\multicolumn{7}{l}{\emph{Workflow-agent baselines}} \\
Workflow Agent (Qwen3.5-9B) & 32.5\% & 40.9\% & 43.6\% & 25.2\% & 10.5\% & 27.9\% \\
Workflow Agent (Gemini-3-Flash) & 47.2\% & 44.6\% & 42.4\% & 57.8\% & 47.4\% & 39.5\% \\
\midrule
\multicolumn{7}{l}{\emph{Qwen3.5-9B training baselines}} \\
SFT~\citep{qwen3.5} & 61.8\% & 59.7\% & 58.2\% & 70.1\% & 66.3\% & 46.5\% \\
GRPO~\citep{shao2024deepseekmath} & 63.5\% & 61.3\% & 60.6\% & 71.4\% & 67.4\% & 48.8\% \\
\texttt{PF-OPSD} & 72.3\% & 71.0\% & 69.7\% & 78.2\% & 75.8\% & 60.5\% \\
\bottomrule
\end{tabular}
\caption{\small Detailed \texttt{VRQABench} evaluation results. Category abbreviations correspond to turn count, turn direction, Sokoban push, direction count, and push-direction count.}
\label{tab:vrqa_detail}
\end{table*}

\begin{table*}[ht]
\centering
\small
\setlength{\tabcolsep}{3pt}
\begin{tabular}{llllllllllllll}
\toprule
Model / Method & Overall & C1 & C2 & C3 & C4 & C5 & C6 & C7 & C8 & C9 & C10 & C11 & C12 \\
\midrule
\multicolumn{14}{l}{\emph{Zero-shot / no-simulation baselines}} \\
\texttt{Gemini-3-Flash}~\citep{pichai2025new} & 48.2 & 61.0 & 35.7 & 61.0 & 35.7 & 65.9 & 40.5 & 71.4 & 50.0 & 21.4 & 53.7 & 40.5 & 42.9 \\
\texttt{GPT-5.4}~\citep{openai2026gpt54} & 53.4 & 58.5 & 19.0 & 65.9 & 50.0 & 68.3 & 57.1 & 71.4 & 47.6 & 28.6 & 65.9 & 54.8 & 54.8 \\
\texttt{HY3}~\citep{tencent2026hy3preview} & 35.0 & 39.0 & 16.7 & 48.8 & 33.3 & 43.9 & 31.0 & 35.7 & 42.9 & 28.6 & 48.8 & 33.3 & 19.0 \\
\texttt{Qwen3.6-27B}~\citep{qwen3.6-27b} & 41.4 & 68.3 & 0.0 & 65.9 & 21.4 & 53.7 & 47.6 & 59.5 & 38.1 & 0.0 & 34.1 & 45.2 & 64.3 \\
\texttt{Qwen3.5-9B}~\citep{qwen3.5} & 39.8 & 68.3 & 0.0 & 68.3 & 2.4 & 36.6 & 54.8 & 57.1 & 35.7 & 0.0 & 31.7 & 52.4 & 71.4 \\
\texttt{Qwen2.5-VL-7B}~\citep{qwen3vl} & 14.2 & 12.2 & 0.0 & 36.6 & 2.4 & 48.8 & 2.4 & 23.8 & 11.9 & 0.0 & 19.5 & 4.8 & 9.5 \\
\midrule
\multicolumn{14}{l}{\emph{Workflow-agent baselines}} \\
Workflow Agent (Qwen3.5-9B) & 38.6 & 65.9 & 0.0 & 65.9 & 2.4 & 36.6 & 52.4 & 57.1 & 33.3 & 0.0 & 31.7 & 50.0 & 69.0 \\
Workflow Agent (Gemini-3-Flash) & 49.6 & 63.4 & 35.7 & 61.0 & 35.7 & 68.3 & 42.9 & 71.4 & 52.4 & 21.4 & 56.1 & 42.9 & 45.2 \\
\midrule
\multicolumn{14}{l}{\emph{Qwen3.5-9B training baselines}} \\
SFT~\citep{qwen3.5} & 59.6 & 75.6 & 31.0 & 78.0 & 40.5 & 61.0 & 69.0 & 71.4 & 61.9 & 19.0 & 68.3 & 57.1 & 83.3 \\
GRPO~\citep{shao2024deepseekmath} & 61.2 & 75.6 & 33.3 & 78.0 & 42.9 & 63.4 & 71.4 & 73.8 & 64.3 & 21.4 & 68.3 & 59.5 & 83.3 \\
\texttt{PF-OPSD} & 70.4 & 80.5 & 47.6 & 85.4 & 54.8 & 70.7 & 78.6 & 81.0 & 73.8 & 38.1 & 78.0 & 66.7 & 90.5 \\
\bottomrule
\end{tabular}
\caption{\small Detailed \texttt{OpenWorldQA} test results. Values are percentages; category names follow the dataset taxonomy.}
\label{tab:openworld_detail}
\end{table*}

\subsection{OpenWorldQA Macro and Resource-Grouped Results}
\label{sec:openworld_macro_resource}

Because the \texttt{OpenWorldQA} training split is intentionally collected from natural videos and is not category-balanced, we additionally report macro-averaged and resource-grouped results. Macro averaging gives equal weight to the 12 physical categories. We also group categories by the number of training examples: low-resource categories have fewer than 100 training examples (C1, C3, C5, C7, C10, C12), medium-resource categories have 100--800 examples (C6, C8, C11), and high-resource categories have more than 800 examples (C2, C4, C9). The test split remains approximately balanced, with 41--42 questions per category.

\begin{table}[ht]
\centering
\small
\begin{tabular}{lcccc}
\toprule
Model / Method & Macro Avg. & Low-resource & Medium-resource & High-resource \\
\midrule
SFT~\citep{qwen3.5} & 59.7 & 72.9 & 62.7 & 30.2 \\
GRPO~\citep{shao2024deepseekmath} & 61.3 & 73.7 & 65.1 & 32.5 \\
\texttt{PF-OPSD} & 70.5 & 81.0 & 73.0 & 46.8 \\
\midrule
\texttt{PF-OPSD} $-$ SFT & +10.8 & +8.1 & +10.3 & +16.6 \\
\bottomrule
\end{tabular}
\caption{\small Macro-averaged and resource-grouped \texttt{OpenWorldQA} results. Values are percentages. Resource groups are defined by the number of training examples per category, while evaluation remains balanced across categories.}
\label{tab:openworld_macro_resource}
\end{table}

PF-OPSD improves over SFT in all resource groups, including categories with fewer than 100 training examples. This suggests that the gains are not only a consequence of repeated motifs in the largest training categories. The high-resource group remains the hardest because it contains spatial relation, support stability, and tool-use questions with visually subtle outcome differences; nevertheless, PF-OPSD gives the largest absolute gain there, indicating that simulation control is especially helpful when static cues and generated rollouts are easy to confuse.

\subsection{Retry Depth Diagnostic}
\label{sec:retry_depth_diagnostic}

Table~\ref{tab:retry_depth_diagnostic} reports the same realized simulation-depth statistics as Figure~\ref{fig:call_depth}. The number of attempts is endogenous: examples with two or three attempts are those for which earlier rollouts are rejected or uncertain. Their lower accuracy should therefore be interpreted primarily as evidence that the policy routes harder examples to deeper retry paths, not as evidence that retry itself causes errors.

\begin{table}[ht]
\centering
\small
\begin{tabular}{lcccc}
\toprule
Attempts & Share & VRQA Acc. & OWQA Acc. & Misjudge \\
\midrule
0 & 57.5\% & 67.2\% & 66.6\% & -- \\
1 & 40.5\% & 78.8\% & 76.9\% & 3.9\% \\
2 & 1.5\% & 71.4\% & 66.7\% & 5.9\% \\
3 & 0.5\% & 66.7\% & 66.7\% & 16.7\% \\
\bottomrule
\end{tabular}
\caption{\small Realized retry-depth diagnostic under PF-OPSD. Attempt depth is selected by the policy and is therefore correlated with sample difficulty.}
\label{tab:retry_depth_diagnostic}
\end{table}

\section{PF-OPSD Algorithm}
\label{sec:pfopsd_algorithm}

Algorithm~\ref{alg:pfopsd_appendix} summarizes the training procedure used to construct the advantage-weighted targets in the main text.

\begin{algorithm}[ht]
\small
\caption{\small Privileged-Future On-Policy Self-Distillation (PF-OPSD)}
\label{alg:pfopsd_appendix}
\begin{algorithmic}[1]
\Require Training sample $(x,y^*,v^*)$ with $x=(o,q,\mathcal{O})$, world model $W$, student policy $\pi_\theta$, privileged evaluator $E^+$, simulation cap $B=3$
\Ensure Updated student policy $\pi_\theta$
\For{each training step}
    \State Generate a student-view trajectory $\tau_s\sim p_\theta(\cdot\mid x;W)$ with $v^*$ hidden
    \For{each decision node $t$ visited in $\tau_s$}
        \If{$t\in\mathcal{C}_{\mathrm{disc}}$}
            \State Set $C_t$ to the valid discrete action set
        \Else
            \State Sample $C_t=\{a_1,\ldots,a_K\}$ from $\pi_\theta(\cdot\mid c_t,h_t^s)$
        \EndIf
        \For{each candidate $a\in C_t$}
            \State Force $a$ at node $t$ and greedily complete $\tau_a$ under the same student-view rollout protocol
            \State Score $R^+(\tau_a;y^*,v^*)$ with $E^+$ using teacher-side context $(y^*,v^*)$
            \State Set $Q_t^+(a)=R^+(\tau_a;y^*,v^*)$
        \EndFor
        \State Compute $V_t^+$ and $A_t^+(a)=Q_t^+(a)-V_t^+$
        \State Construct discrete targets $q_t^\star$ or text weights $w_{t,k}$
    \EndFor
    \State Update $\pi_\theta$ with $\mathcal{L}_{\mathrm{disc}}+\mathcal{L}_{\mathrm{text}}$
\EndFor
\end{algorithmic}
\end{algorithm}

\section{Related Work}
\label{sec:appendix_related_work}

\subsection{Multimodal and Tool-Augmented Reasoning}
Large language models have shown strong language-based reasoning abilities through structured prompting and intermediate inference, including chain-of-thought prompting, self-consistency, and zero-shot reasoning \citep{wei2022chain,wang2022self,KojimaGRMI22,zhou2023thread,tao2024magis}. Subsequent work extends this reasoning process with external computation and interaction: program-aided reasoning separates symbolic reasoning from executable calculation \citep{chen2022program}, while ReAct and Toolformer demonstrate that language models can interleave reasoning with tool invocation and observations \citep{yao2023react,schick2023toolformer}. These lines of work establish the importance of external resources, but most tools return relatively explicit textual, symbolic, or numerical outputs.

Multimodal large language models further connect language reasoning with visual perception, enabling image-text dialogue, visual question answering, and complex multimodal instruction following \citep{SurveyMLLM,VisualMLLM,qwen3vl,zhou2024visual}. Nevertheless, future-oriented reasoning remains difficult because the required future state may not be visible in the input image. The model must infer motion tendencies, spatial constraints, contact changes, and causal consequences from a static state. This exposes the gap between language-centric abstraction and grounded dynamic prediction, echoing broader observations that language-only or weakly grounded models can lack stable experiential grounding \citep{bisk2020experience,mccoy2023language}.

\subsection{Visual Reasoning, Physical Prediction, and Benchmarks}
Visual reasoning has long studied how models combine perception with compositional structure and physical regularities. Diagnostic benchmarks such as CLEVR and GQA evaluate compositional visual question answering \citep{johnson2017clevr,hudson2019gqa}, while physical reasoning benchmarks such as PHYRE and IntPhys focus on intuitive physics and outcome prediction \citep{bakhtin2019phyre,riemer2020learning}. These benchmarks inspired architectures for relational and compositional reasoning, including relational networks, FiLM, and compositional attention \citep{santoro2017simple,perez2018film,hudson2018compositional}. More recent self-supervised visual representations also capture useful structural regularities from large-scale image data \citep{oquab2023dinov2,assran2023self}.

Despite this progress, many existing visual reasoning tasks either emphasize static recognition or evaluate perception from observed videos. Our setting is different: the agent receives only a current image or pre-event anchor frame and must answer a question about a future outcome. This makes the task closer to foresight for embodied and interactive agents. \texttt{VRQABench} builds on VR-Bench \citep{VRBench} to test controllable spatial lookahead in maze and Sokoban-style environments, while \texttt{OpenWorldQA} complements it with open-domain physical prediction from real-world anchor frames. Together, they focus on whether a model can infer future states from static initial conditions rather than merely recognize completed events.

\subsection{Generative World Models as Concrete-Reasoning Sources}
Video representation learning and video generation have expanded visual reasoning from static perception to temporal dynamics. Predictive video representation methods and masked video modeling learn temporal regularities from video data \citep{tong2022videomae,bardes2024vjepa}, and recent studies suggest that video models can exhibit zero-shot physical and spatial reasoning through frame-by-frame generation \citep{wiedemer2025video}. At the same time, large-scale generative video models are increasingly viewed as world models that can synthesize plausible future rollouts conditioned on an initial image and text instruction \citep{wan2025,HunyuanVideo,Simulatingwm,Physicalwm}.

{Recent work also uses generated or interleaved visual content as an explicit reasoning medium. MVoT generates intermediate visualizations for spatial reasoning \citep{li2025mvot}; MindJourney scales spatial inference through world-model exploration \citep{yang2025mindjourney}; ThinkMorph interleaves textual and visual chains of thought \citep{gu2025thinkmorph}; and multimodal-world-model studies examine when visual generation supports human-like reasoning \citep{wu2026visualgeneration}. Our contribution is not the first use of generated visual content for reasoning. Instead, we focus on future-oriented visual QA in which external rollouts are optional and fallible, and learn when to invoke them, whether to trust them, and how to arbitrate them against abstract reasoning.}

Using such models as concrete-reasoning sources is attractive for future prediction because generated rollouts may expose object motion, path evolution, contact changes, and temporal consequences that are absent from the static input. Recent agent work has explored world models in structured digital environments, such as GUI transition simulation for app agents \citep{luo2025vimo,cao2026mobiledreamer}. However, world models are not precise oracles. They may hallucinate visually plausible but task-incorrect futures, alter important geometry, ignore small causal factors, or generate rollouts with uncertain task utility. Recent evidence further suggests that current agents often fail to leverage world models effectively for foresight \citep{qian2026current}. Therefore, world-model assistance should be treated as noisy concrete reasoning rather than unconditional truth.

\subsection{Learning to Control Concrete Reasoning}
The above observations motivate a shift from simple world-model attachment to controlled concrete reasoning. A generative world model differs from conventional tools because it returns high-dimensional, continuous, and potentially misleading visual rollouts. An MLLM must therefore decide whether simulation is necessary, formulate an appropriate query, verify whether the generated rollout is credible, and resolve conflicts between abstract reasoning and rollout-based concrete reasoning. This makes world-model-assisted reasoning a problem of simulation selection, rollout verification, and rollout reliance rather than only tool invocation.

Our PF-OPSD framework is related to knowledge distillation, learning with privileged information, and on-policy learning \citep{hinton2015distilling,vapnik2009new,schulman2017proximal}. Standard distillation transfers teacher supervision into a deployable student, while privileged-information settings use training-only signals that are unavailable at test time. PF-OPSD should be viewed as an instantiation of these ideas for world-model-assisted future QA rather than as a new general-purpose distillation principle. The specific contribution is to make the privileged signal act on simulation-control nodes: whether to call a world model, whether to accept or reject a rollout, how to rely on it, and which final answer to produce. In our case, ground-truth future videos provide privileged context for calibrating whether rollouts and intermediate concrete-reasoning decisions are useful. Unlike offline imitation of fixed demonstrations, the policy first samples its own trajectories under test-time inputs, and the privileged future is used only to produce soft targets for decision nodes visited by the current policy. This on-policy design aligns training with the states the deployable agent actually encounters, while keeping true futures unavailable during inference.

\section{Dataset Details}
\label{sec:dataset_details}

This section provides the full category definitions and split distributions for the two benchmarks summarized in the main paper.

\subsection{Benchmark Verification Protocol}
\label{sec:benchmark_verification_protocol}

Both benchmarks are manually verified after automatic filtering. Human verification is used as a final quality gate rather than as a source of model input. Retained examples must have a valid initial observation, a unique answer, plausible distractors, and no leakage of the future outcome. Ambiguous, visually inconsistent, or option-invalid items are removed.

\begin{table}[ht]
\centering
\small
\setlength{\tabcolsep}{4pt}
\begin{tabular}{p{0.13\linewidth}p{0.40\linewidth}p{0.40\linewidth}}
\toprule
Dataset & Human verification checks & Removed case types \\
\midrule
\texttt{VRQABench} & Visual consistency; unique solver-derived answer; plausible options & Invalid question text; implausible option; ambiguous puzzle state \\
\texttt{OpenWorldQA} & Anchor validity; answer uniqueness; no future leakage; option plausibility & Ambiguous outcome; leaked anchor; visually inconsistent question; invalid option \\
\bottomrule
\end{tabular}
\caption{\small Minimal verification protocol for the two benchmark suites.}
\label{tab:benchmark_verification_protocol}
\end{table}

\subsection{\texttt{VRQABench} Category Taxonomy and Distribution}

\texttt{VRQABench} is built from three VR-Bench task families: 2D maze navigation, irregular-maze path tracing, and Sokoban box pushing. Each example is a four-choice question generated from an initial puzzle image. Table~\ref{tab:vrqa_category_definitions} defines the five retained spatial question categories, Table~\ref{tab:vrqa_distribution} reports their train/evaluation distributions, and Table~\ref{tab:vrqa_task_distribution} summarizes the task-family distribution.

\begin{table*}[ht]
\centering
\small
\begin{tabular}{lll}
\toprule
Category & Applicable tasks & Required reasoning \\
\midrule
\texttt{C1\_turn\_count} & maze, irregular maze & Count total turns along the optimal path. \\
\texttt{C2\_turn\_direction} & maze, irregular maze & Count left or right turns along the optimal path. \\
\texttt{C3\_sokoban\_push} & Sokoban & Infer minimum push count or the first push direction. \\
\texttt{C4\_direction\_count} & maze & Count optimal-path steps in a specified direction. \\
\texttt{C5\_push\_dir\_count} & Sokoban & Count pushes in a specified direction in the optimal solution. \\
\bottomrule
\end{tabular}
\caption{\small \texttt{VRQABench} category definitions.}
\label{tab:vrqa_category_definitions}
\end{table*}

\begin{table}[ht]
\centering
\small
\begin{tabular}{lrrr}
\toprule
Category & Train & Eval & Total \\
\midrule
\texttt{C1\_turn\_count} & 1,194 & 186 & 1,380 \\
\texttt{C2\_turn\_direction} & 1,033 & 165 & 1,198 \\
\texttt{C3\_sokoban\_push} & 928 & 147 & 1,075 \\
\texttt{C4\_direction\_count} & 637 & 95 & 732 \\
\texttt{C5\_push\_dir\_count} & 208 & 43 & 251 \\
\midrule
Total & 4,000 & 636 & 4,636 \\
\bottomrule
\end{tabular}
\caption{\small \texttt{VRQABench} category distribution across train and evaluation splits.}
\label{tab:vrqa_distribution}
\end{table}

\begin{table}[ht]
\centering
\small
\begin{tabular}{lrrr}
\toprule
Task type & Train & Eval & Total \\
\midrule
maze & 1,756 & 275 & 2,031 \\
irregular maze & 1,108 & 171 & 1,279 \\
Sokoban & 1,136 & 190 & 1,326 \\
\midrule
Total & 4,000 & 636 & 4,636 \\
\bottomrule
\end{tabular}
\caption{\small \texttt{VRQABench} task-family distribution.}
\label{tab:vrqa_task_distribution}
\end{table}

\subsection{\texttt{OpenWorldQA} Category Taxonomy and Distribution}

\texttt{OpenWorldQA} contains four-choice questions from real-world short videos. Each question uses an anchor frame before the outcome is visible and asks the model to predict a future physical event. Table~\ref{tab:openworld_category_definitions} defines the 12 physical reasoning categories, Table~\ref{tab:openworld_category_distribution} reports the train/test category distribution, and Table~\ref{tab:openworld_question_type_distribution} reports the question-type distribution.

\begin{table*}[ht]
\centering
\small
\setlength{\tabcolsep}{4pt}
\begin{tabular}{llp{0.58\linewidth}}
\toprule
Category & Name & Description \\
\midrule
\texttt{C1\_fit\_clearance} & fit clearance & Infer the intermediate pose or orientation needed for a rigid object to pass through limited clearance. \\
\texttt{C2\_spatial} & spatial relation & Predict final position or region after a physical event. \\
\texttt{C3\_containment} & containment & Reason about overflow, filling, liquid distribution, or whether a container can hold the material. \\
\texttt{C4\_support} & support stability & Predict stability, tipping direction, collapse, or balance. \\
\texttt{C5\_friction} & friction & Predict sliding distance, slipping versus gripping, or stopping position. \\
\texttt{C6\_inertia} & inertia & Predict trajectory, direction, or distance after force application or removal. \\
\texttt{C7\_fluidity} & fluidity & Predict flow path, spill threshold, splash pattern, or absorption. \\
\texttt{C8\_deformability} & deformability & Predict stretching, tearing, wrinkling, breaking points, or elastic versus permanent deformation. \\
\texttt{C9\_tool\_use} & tool use & Predict the physical result of a tool acting on an object. \\
\texttt{C10\_chain\_reaction} & chain reaction & Predict indirect or downstream effects beyond the first contact point. \\
\texttt{C11\_process\_race} & process race & Decide which of two concurrent physical processes completes first. \\
\texttt{C12\_multi\_body} & multi-body motion & Reason about momentum transfer, competing forces, or spatial conflicts among multiple bodies. \\
\bottomrule
\end{tabular}
\caption{\small \texttt{OpenWorldQA} category definitions.}
\label{tab:openworld_category_definitions}
\end{table*}

\begin{table*}[ht]
\centering
\small
\setlength{\tabcolsep}{10pt}
\begin{tabular}{lrrrrr}
\toprule
Category & Train & Train \% & Test & Test \% & Total \\
\midrule
\texttt{C1\_fit\_clearance} & 34 & 45.3 & 41 & 54.7 & 75 \\
\texttt{C2\_spatial} & 1,062 & 96.2 & 42 & 3.8 & 1,104 \\
\texttt{C3\_containment} & 31 & 43.1 & 41 & 56.9 & 72 \\
\texttt{C4\_support} & 843 & 95.3 & 42 & 4.7 & 885 \\
\texttt{C5\_friction} & 63 & 60.6 & 41 & 39.4 & 104 \\
\texttt{C6\_inertia} & 168 & 80.0 & 42 & 20.0 & 210 \\
\texttt{C7\_fluidity} & 87 & 67.4 & 42 & 32.6 & 129 \\
\texttt{C8\_deformability} & 443 & 91.3 & 42 & 8.7 & 485 \\
\texttt{C9\_tool\_use} & 853 & 95.3 & 42 & 4.7 & 895 \\
\texttt{C10\_chain\_reaction} & 33 & 44.6 & 41 & 55.4 & 74 \\
\texttt{C11\_process\_race} & 200 & 82.6 & 42 & 17.4 & 242 \\
\texttt{C12\_multi\_body} & 87 & 67.4 & 42 & 32.6 & 129 \\
\midrule
Total & 3,904 & 88.6 & 500 & 11.4 & 4,404 \\
\bottomrule
\end{tabular}
\caption{\small \texttt{OpenWorldQA} category distribution across train and test splits. The test split is approximately balanced across categories, with 41--42 questions per category.}
\label{tab:openworld_category_distribution}
\end{table*}

\begin{table}[ht]
\centering
\small
\begin{tabular}{lrr}
\toprule
Question type & Count & Share \\
\midrule
\texttt{[ORDER]} & 1,458 & 33.1 \\
\texttt{[FIRST-CONTACT]} & 1,213 & 27.5 \\
\texttt{[INTERMEDIATE]} & 1,093 & 24.8 \\
\texttt{[COUNT]} & 254 & 5.8 \\
\texttt{[COUNTERFACTUAL]} & 208 & 4.7 \\
\texttt{[FAILURE]} & 178 & 4.0 \\
\midrule
Total & 4,404 & 100.0 \\
\bottomrule
\end{tabular}
\caption{\small \texttt{OpenWorldQA} question-type distribution.}
\label{tab:openworld_question_type_distribution}
\end{table}

\section{Workflow-Agent Prompt Template}
\label{sec:workflow_agent_prompt}

For the workflow-agent baselines, we use a fixed prompt that exposes Helios as an optional world-model tool while leaving all simulation decisions to the base MLLM. The agent receives only the initial image or anchor frame, the question, and the answer options; it never receives the ground-truth future or answer. A Helios query returns a 100-frame rollout, and the agent may issue at most three simulation queries per example.

\begin{tcblisting}{
colback=white,
colframe=black,
arc=1mm,
auto outer arc,
title={Workflow-Agent System Prompt},
breakable,
listing only,
listing options={basicstyle=\small\ttfamily, breaklines=true, columns=fullflexible, keepspaces=true, showstringspaces=false}
}
You are a multimodal future-prediction agent equipped with an optional video world-model tool. Your goal is to answer four-choice questions about what will happen after the current image or anchor frame.

\textbf{Inputs.} You are given an initial observation, a question, and four answer options. The future video and the correct answer are hidden from you.

\textbf{Capabilities.}
\begin{enumerate}[leftmargin=*, itemsep=1pt, topsep=1pt]
    \item \textbf{Abstract reasoning.} You can reason from the static visual evidence, question, options, spatial constraints, physical rules, and task context.
    \item \textbf{Concrete simulation.} When the future outcome depends on non-trivial motion, contact, path following, object interaction, or physical dynamics, you may call the world model to generate a candidate future rollout.
\end{enumerate}

\textbf{Protocol.}
\begin{itemize}[leftmargin=*, itemsep=1pt, topsep=1pt]
    \item First decide whether the answer can be inferred reliably from the static observation alone.
    \item If simulation is useful, output \texttt{<SIMULATE>} followed by a concise, task-specific simulation prompt describing the relevant objects, motion, path, contact, or physical event to roll out. The simulator will return a 100-frame generated video.
    \item After receiving a rollout, act as a gatekeeper before using it. If the rollout is visually consistent with the input, physically plausible, and relevant to the question, output \texttt{<ACC>} and incorporate it into your reasoning. If it changes important geometry, misses the queried interaction, hallucinates objects, looks glitchy, or conflicts with physical constraints, output \texttt{<REJ>} and either retry with a better prompt or fall back to abstract reasoning.
    \item You may use at most three simulation attempts. Rejected or uncertain rollouts should be discounted rather than treated as ground truth.
    \item Finish with exactly one choice in the format \texttt{Final Answer: [A/B/C/D]}.
\end{itemize}
\end{tcblisting}

\section{Dataset Construction Prompts}
\label{sec:appendix}

For reproducibility, we include the dataset-construction prompt materials used to build the two benchmarks. Each dataset begins with a short prose overview for context, followed by the actual stage or agent prompts shown verbatim; only non-ASCII decorative symbols such as box-drawing characters, arrows, and warning/check icons are normalized to ASCII for LaTeX compatibility. After these automatic stages, every retained item in both datasets is manually verified by human annotators, and samples with ambiguous answers, invalid anchors, implausible options, or visual inconsistencies are removed.

\subsection{Full \texttt{VRQABench} Dataset-Construction Prompt}
The \texttt{VRQABench} v2 pipeline replaces the original VLM-based solution tracing with a programmatic solver, eliminating hallucination in the answer generation stage. VLM calls are limited to writing question text and reviewing question quality. The overall construction flow is: \texttt{state.json} $\rightarrow$ programmatic solver $\rightarrow$ \texttt{build\_qa\_items()} $\rightarrow$ QuestionWriter $\rightarrow$ SmallModelProbe $\rightarrow$ Reviewer $\rightarrow$ human verification $\rightarrow$ \texttt{output/reviewed/}. The actual stage prompts and specifications are shown below.

\begin{tcblisting}{
colback=white,
colframe=black,
arc=1mm,
auto outer arc,
title={VRQABench Step 1: Programmatic Solver},
breakable,
listing only,
listing options={basicstyle=\small\ttfamily, breaklines=true, columns=fullflexible, keepspaces=true, showstringspaces=false}
}
## Stage 1 - Programmatic Solver

**Role.** Reads `state.json` (ground-truth game state bundled with each puzzle) and computes the exact solution via search or geometry. No VLM involved.

**Solver by task type.**

| Task | Algorithm | Key outputs |
|------|-----------|-------------|
| maze | BFS on grid (`grid.data`, walls = 1) | `steps`, `total_turns`, `left_turns`, `right_turns` |
| irregular\_maze | Geometric angle analysis on `solution_path` waypoints (turn threshold 30 degrees) | `direction_changes`, `left_turns`, `right_turns` |
| sokoban | Push-level BFS over `(player_pos, box_positions)` states with inner reachability check | `push_sequence`, `total_pushes`, `first_push_direction` |

**Turn counting rules.**
- maze: a turn is any step where the movement direction differs from the previous step; a U-turn (180 degrees) counts as two turns.
- irregular\_maze: a junction is a turn when the angle between incoming and outgoing vectors exceeds 30 degrees; straight-through junctions are ignored.

**Direction conventions.**
- maze / sokoban: `up`, `down`, `left`, `right` (image coordinate frame, y-axis down).
- irregular\_maze: turns are classified as left or right using the cross product of consecutive direction vectors (image coordinate frame: cross > 0 -> right).
\end{tcblisting}

\begin{tcblisting}{
colback=white,
colframe=black,
arc=1mm,
auto outer arc,
title={VRQABench Step 2: QA Item Builder},
breakable,
listing only,
listing options={basicstyle=\small\ttfamily, breaklines=true, columns=fullflexible, keepspaces=true, showstringspaces=false}
}
## Stage 2 - build_qa_items()

**Role.** Programmatically generates answer values and distractor options for each applicable question category. No VLM. The `question` field is left empty at this stage.

**Question categories.**

| Category | Question asked | Applicable tasks |
|---------|------------|--------------|
| C1\_turn\_count | How many times does the path change direction in the optimal solution? | maze, irregular\_maze |
| C2\_turn\_direction | How many times does the path turn left (or right) in the optimal solution? | maze, irregular\_maze |
| C3\_sokoban\_push | (a) What is the minimum number of pushes to solve the puzzle? (b) In which direction is the first push? | sokoban |
| C4\_direction\_count | How many steps does the path take going \<direction\> in the optimal solution? | maze |
| C5\_push\_dir\_count | How many times is the box pushed \<direction\> in the optimal solution? | sokoban |

For C4 and C5, the direction asked is the **most-frequent direction** in the solution (to avoid trivially-zero answers).

**Questions generated per sample.**

| Task | Questions |
|------|-----------|
| maze | up to 3 (C1 + C2 + C4) |
| irregular\_maze | up to 2 (C1 + C2) |
| sokoban | up to 3 (C3-count + C3-direction + C5) |

**Distractor design.**
- *Counting questions*: distractors are sampled from integers within +/-4 of the correct value (all >= 0); correct answer placed at a randomly chosen option letter.
- *Direction questions*: all four cardinal directions appear exactly once across the four options; correct answer placed at a random letter.
\end{tcblisting}

\begin{tcblisting}{
colback=white,
colframe=black,
arc=1mm,
auto outer arc,
title={VRQABench Step 3: QuestionWriter Prompt},
breakable,
listing only,
listing options={basicstyle=\small\ttfamily, breaklines=true, columns=fullflexible, keepspaces=true, showstringspaces=false}
}
## Stage 3 - QuestionWriter (VLM)

**Role.** Receives the puzzle image and the category/direction metadata. Writes ONE natural-language question sentence describing the puzzle visually. Does **not** generate or modify the answer.

**Model.** gpt-5.5 (MODEL_LARGE).

**Output format.**
```json
{"question": "Starting from the ball in the upper-right corner, following the shortest path to the green square at the bottom-left, how many times does the path change direction?"}
```

**Writing constraints.**
- Mention start and goal positions using visible landmarks (not grid coordinates).
- Use the exact question phrasing for the category.
- Do **not** include any numbers, directions, or hints toward the correct answer.
- One sentence is ideal, two at most.
\end{tcblisting}

\begin{tcblisting}{
colback=white,
colframe=black,
arc=1mm,
auto outer arc,
title={VRQABench Step 4: SmallModelProbe Filter},
breakable,
listing only,
listing options={basicstyle=\small\ttfamily, breaklines=true, columns=fullflexible, keepspaces=true, showstringspaces=false}
}
## Pre-Reviewer Filter - SmallModelProbe

Before entering the Reviewer stage, each candidate question is subjected to a difficulty probe using a smaller model (GPT-5.4-nano):

1. The question is presented **twice**, each time with the option order independently shuffled.
2. If the probe answers correctly on **both** attempts, the question is classified as `too_easy` and discarded without reaching the Reviewer.

This filter removes questions that can be answered correctly without genuine spatial reasoning.
\end{tcblisting}

\begin{tcblisting}{
colback=white,
colframe=black,
arc=1mm,
auto outer arc,
title={VRQABench Step 5: Reviewer Prompt},
breakable,
listing only,
listing options={basicstyle=\small\ttfamily, breaklines=true, columns=fullflexible, keepspaces=true, showstringspaces=false}
}
## Stage 4 - Reviewer (VLM)

**Role.** Receives the puzzle image and the QA list. Evaluates **question text quality only** - the answer key is trusted as correct (programmatically verified) and is not re-examined.

**Model.** gpt-5.5 (MODEL_LARGE).

**Verification criteria.**

| Criterion | Pass condition |
|-----------|---------------|
| Question text validity | The question text accurately describes features visible in the puzzle image (start/goal positions, task type language). |
| Distractor plausibility | Each wrong option is plausible; no option is obviously impossible given the visible puzzle layout. |
| Difficulty | The answer requires genuine path-tracing reasoning; it cannot be read off the image trivially. |

**Acceptance rule.** A question is accepted if and only if all three criteria pass and the overall quality score is >= 7 out of 10.
\end{tcblisting}

\begin{tcblisting}{
colback=white,
colframe=black,
arc=1mm,
auto outer arc,
title={VRQABench Step 6: Final Data Format},
breakable,
listing only,
listing options={basicstyle=\small\ttfamily, breaklines=true, columns=fullflexible, keepspaces=true, showstringspaces=false}
}
## Final Data Format

Each accepted question is saved as an individual JSON file (`VRB_{category}_{NNNN}.json`):

```json
{
  "category": "C1_turn_count",
  "question": "Starting from the ball in the upper-right corner, following the shortest path to the green square at the bottom-left, how many times does the path change direction?",
  "options": {"A": "8", "B": "11", "C": "9", "D": "10"},
  "answer": "D",
  "correct_value": "10",
  "source_sample_id": "maze_1_hard_0097_0",
  "task_type": "maze",
  "input_image": "<absolute path to puzzle image>",
  "video_path": "<absolute path to solution video>",
  "_solution": { "total_turns": 10, "left_turns": 5, "right_turns": 5, "..." : "..." },
  "_review": { "decision": "accept", "score": 9, "..." : "..." },
  "_pipeline_metadata": { "pipeline_version": "v2", "..." : "..." }
}
```
\end{tcblisting}

\subsection{Full \texttt{OpenWorldQA} Dataset-Construction Prompts}
The \texttt{OpenWorldQA} dataset is constructed with a five-stage multi-agent pipeline: SceneAnalyst, QuestionDesigner, DistractorForge, SmallModelProbe, and Reviewer. The pipeline starts from a video frame sequence, produces a structured scene report and anchor frame, designs question skeletons, fills plausible distractors, filters questions with two small-model probes, and applies a five-dimensional reviewer. The accepted items are then manually verified by human annotators before being written to the train/test splits. The actual agent prompts are shown below.

\begin{tcblisting}{
colback=white,
colframe=black,
arc=1mm,
auto outer arc,
title={OpenWorldQA Step 1: SceneAnalyst Prompt},
breakable,
listing only,
listing options={basicstyle=\small\ttfamily, breaklines=true, columns=fullflexible, keepspaces=true, showstringspaces=false}
}
## Agent 1: SceneAnalyst

**Role:** Receives the full frame sequence, produces a structured physical scene report, and selects the best anchor frame.

---

```
You are SceneAnalyst - the first agent in a multi-stage physical reasoning QA pipeline.

You receive a sequence of frames extracted from a short video clip (up to 12 frames, ~0.5s apart,
covering the first 10 seconds of the video). Your job is twofold:

  1. Produce a structured physical scene report that later agents will use to design hard questions.
  2. Select the best anchor frame - the single frame that best represents the moment just BEFORE
     the main physical event reaches its outcome (i.e., initial-condition frame).

==============================
PART 1 - SCENE REPORT
==============================

Analyse ALL provided frames carefully and fill every section below.

[OBJECTS]
List every physically relevant object visible. For each object state:
  - name, approximate size (small/medium/large relative to scene), material or texture cues,
    deformability (rigid / flexible / liquid), and current state (static / moving / being held).

[AGENTS]
Describe the people present: number, body position (standing/sitting/crouching),
which hand(s) are active, and what they appear to be doing.

[FORCES & MOTION]
Identify any visible forces or motions across the frame sequence:
  - direction of movement (toward / away / left / right / downward / rotating)
  - estimated speed (slow / moderate / fast)
  - contact points between objects or between person and object
  - any visible momentum, spin, tilt, or instability

[PHYSICAL STATE AT EACH FRAME]
For each frame (frame_001 ... frame_N), write one sentence describing the key physical state
change compared to the previous frame. Focus on: position change, contact change, deformation,
fluid level change, or stability change.

[CANDIDATE EVENTS]
List up to 3 distinct physical events that are either in progress or clearly about to happen.
For each event describe:
  - what is happening physically
  - what the uncertain outcome is (i.e., what a model must PREDICT, not just observe)
  - which frames bracket the event (e.g., "frame_003 to frame_007")

[SUITABLE CATEGORIES]
From the list below, identify which categories are NATURALLY AND CLEARLY present in this clip.
Only include a category if there is genuine visual evidence for it - do not guess.

  C1_fit_clearance      - predicting whether an object fits through/into a space given size cues
  C2_spatial            - predicting final spatial position/region after a physical event
  C3_containment        - overflow, underfill, fit vs. no-fit, liquid distribution
  C4_support            - stability, tipping direction, collapse, balance
  C5_friction           - sliding distance, slip vs. grip, stopping point
  C6_inertia            - trajectory, momentum after release, spin, rebound direction
  C7_fluidity           - flow path, overflow threshold, splash pattern, absorption
  C8_deformability      - stretch, tear, crumple, snap, permanent vs. elastic deformation
  C9_tool_use           - physical outcome of tool acting on target (not human intent)
  C10_chain_reaction    - indirect / downstream effect beyond the direct contact
  C11_process_race      - which of two concurrent physical processes completes first
  C12_multi_body        - momentum transfer, force contest, or spatial conflict between bodies

==============================
PART 2 - ANCHOR FRAME SELECTION
==============================

Select the single best anchor frame using this criterion:

  The anchor must show the INITIAL CONDITIONS of the main physical event - enough context
  for a reasoning model to predict the outcome - but must NOT yet reveal the outcome itself.

Rules:
  - Prefer the frame just BEFORE the critical transition (e.g., object about to be released,
    liquid about to reach the rim, object tipping but not yet fallen).
  - Reject frames where the answer is already visible (object already fallen, already spilled).
  - Reject the very first frame if it appears to be a cover/title/static shot with no action.
  - If the video has no clear physical event, still pick the most informative frame.

==============================
OUTPUT FORMAT
==============================

Return a single JSON object with the following structure (ALL text in English):

{
  "objects": [
    {"name": "...", "size": "...", "material": "...", "deformability": "...", "state": "..."}
  ],
  "agents": "...",
  "forces_and_motion": "...",
  "frame_states": {
    "frame_001": "...",
    "frame_002": "...",
    ...
  },
  "candidate_events": [
    {
      "description": "...",
      "uncertain_outcome": "...",
      "frame_range": "frame_XXX to frame_YYY"
    }
  ],
  "suitable_categories": ["C1_fit_clearance", "C4_support", ...],
  "anchor_frame": "frame_XXX",
  "anchor_rationale": "One sentence explaining why this frame is the best anchor."
}

Output the JSON and stop. Do not add commentary outside the JSON.
```
\end{tcblisting}

\begin{tcblisting}{
colback=white,
colframe=black,
arc=1mm,
auto outer arc,
title={OpenWorldQA Step 2: QuestionDesigner Prompt},
breakable,
listing only,
listing options={basicstyle=\small\ttfamily, breaklines=true, columns=fullflexible, keepspaces=true, showstringspaces=false}
}
## Agent 2: QuestionDesigner

**Role:** Receives the scene report (text only), designs 6 question skeletons without answer options - those are filled in by DistractorForge.

---

```
You are QuestionDesigner - the second agent in a multi-stage physical reasoning QA pipeline.

You receive a structured scene report produced by SceneAnalyst (text only - no images).
Your job is to design 6 candidate question SKELETONS that will become hard physical-reasoning
multiple-choice questions. You do NOT write the answer options yet - that is DistractorForge's job.

==============================
CORE PRINCIPLE
==============================

A question is only hard if the answer REQUIRES mentally simulating 1-3 future steps.

Self-check before finalising each question:
  "If I did NOT run a short mental simulation of what happens next,
   could I still answer this correctly more than 60% of the time?"
  If YES -> discard this question and try a different type.

==============================
QUESTION TYPES - you MUST use only these six
==============================

For each question, assign exactly one type tag:

  [ORDER]           Which of two specific events happens FIRST from the current state?
  [COUNT]           How many times will X change / switch / contact before the action ends?
  [FIRST-CONTACT]   What does the subject first make contact with next?
  [INTERMEDIATE]    What transient/transitional state does the system enter BEFORE the final outcome?
  [FAILURE]         If uncorrected, which specific step or component fails first?
  [COUNTERFACTUAL]  If the current posture / force / direction is maintained unchanged,
                    which branch does the future trajectory approach?

==============================
STRICTLY FORBIDDEN question patterns
==============================

Do NOT generate questions of these forms - they rely on high scene-prior and do not require
future simulation:

  [REJECT] "What will the person do next?"
  [REJECT] "Will X succeed or fail?"
  [REJECT] "What is the person trying to do?"
  [REJECT] "Will the object be picked up / opened / poured / placed?"
  [REJECT] Any question whose correct answer follows directly from gravity alone
    ("the object will fall") without requiring specific scene details.

==============================
CATEGORY ALIGNMENT
==============================

The scene report includes "suitable_categories". Assign each question to one of those categories.
Try to cover at least 3 different categories across your 6 skeletons.

Category definitions for reference:
  C1_fit_clearance   - fit/gap prediction based on visible size relationships
  C2_spatial         - final resting position or region after motion
  C3_containment     - overflow, underfill, fit vs. no-fit
  C4_support         - tipping direction, collapse, stability
  C5_friction        - sliding endpoint, slip vs. grip
  C6_inertia         - trajectory / direction / distance after force applied or removed
  C7_fluidity        - flow path, overflow threshold, splash, absorption
  C8_deformability   - stretch/tear/snap/buckle point
  C9_tool_use        - physical result of tool on target (not human intent)
  C10_chain_reaction - indirect downstream effect beyond direct contact
  C11_process_race   - which concurrent process completes first
  C12_multi_body     - momentum transfer, force contest between bodies

==============================
ANCHOR AWARENESS
==============================

The scene report specifies an anchor frame. Every question MUST be answerable from that frame
alone (given physical reasoning). Do NOT design questions that require information only visible
AFTER the anchor frame - the test model will only see the anchor.

[WARNING]  NEVER reference frame numbers in question text (e.g., do NOT write "From frame_005",
    "At frame_008", "As seen in frame_003", etc.).
    The question must describe the scene state in plain language only.
    The anchor frame identity is handled by the pipeline - not by the question text.

==============================
OUTPUT FORMAT
==============================

Return a single JSON object (ALL text in English):

{
  "question_skeletons": [
    {
      "id": 1,
      "type": "[ORDER] / [COUNT] / [FIRST-CONTACT] / [INTERMEDIATE] / [FAILURE] / [COUNTERFACTUAL]",
      "category": "C_X_name",
      "question_text": "Full question sentence referencing specific objects and scene details.",
      "key_physical_principle": "One sentence: what physical knowledge is needed to answer this?",
      "true_answer_direction": "Brief description of what the correct answer should describe
                                (NOT the exact wording - DistractorForge will write options).",
      "why_hard": "One sentence explaining why this cannot be answered without simulation.",
      "anchor_frame": "frame_XXX"
    },
    ...  (exactly 6 skeletons)
  ]
}

Output the JSON and stop.
```
\end{tcblisting}

\begin{tcblisting}{
colback=white,
colframe=black,
arc=1mm,
auto outer arc,
title={OpenWorldQA Step 3: DistractorForge Prompt},
breakable,
listing only,
listing options={basicstyle=\small\ttfamily, breaklines=true, columns=fullflexible, keepspaces=true, showstringspaces=false}
}
## Agent 3: DistractorForge

**Role:** Receives the scene report and question skeletons, generates complete four-option QA items for each skeleton, and selects the best 3 out of 6.

---

```
You are DistractorForge - the third agent in a multi-stage physical reasoning QA pipeline.

You receive:
  1. The structured scene report from SceneAnalyst (text only).
  2. The 6 question skeletons from QuestionDesigner (text only).

Your job is to turn each skeleton into a complete multiple-choice QA item with 4 options
(A/B/C/D) where ALL four options are physically plausible in the scene context.

==============================
DISTRACTOR QUALITY - this is your primary responsibility
==============================

Every wrong option (distractor) MUST satisfy ALL three conditions:

  1. PHYSICALLY POSSIBLE in the real world under some conditions.
  2. LOCALLY PLAUSIBLE in THIS specific scene - a viewer unfamiliar with the exact physics
     could reasonably believe it might happen given the visible setup.
  3. CLEARLY WRONG when the anchor frame is examined carefully using correct physical reasoning.

ABSOLUTELY FORBIDDEN distractor types (instant reject):
  [REJECT] Objects floating, levitating, or defying gravity
  [REJECT] Objects freezing in midair without cause
  [REJECT] Objects passing through solid surfaces
  [REJECT] Motion reversing spontaneously without an applied force
  [REJECT] Any outcome that violates basic Newtonian physics

GOOD distractor strategies - make options differ in:
  - Degree    : "slides 10cm past the edge" vs "stops just before the edge" vs "stops 5cm short"
  - Direction : "tips left" vs "tips right" vs "slides forward off the surface"
  - Timing    : "overflows immediately" vs "overflows after 2 more pours" vs "never reaches the rim"
  - Part      : "snaps at the midpoint" vs "snaps near the grip end" vs "bends permanently but holds"
  - Sequence  : "A happens before B" vs "B happens before A" vs "they happen simultaneously"

The four options should be difficult to distinguish without careful examination of the anchor.
A careless observer should find at least 2 options tempting.

==============================
QUESTION-ONLY SELF-CHECK
==============================

Before finalising each QA item, mentally cover the scene report and ask:
  "If I only read the question text (no image), could I guess the correct answer
   more than 60% of the time based on world knowledge alone?"

If YES -> the question is too high-prior. Rewrite the question to reference specific
scene details that anchor it (e.g., "given the visible tilt angle of ~30 degrees",
"given the surface appears to be smooth tile", "given the cup is already ~80% full").

==============================
QUESTION TEXT RULES
==============================

[WARNING]  NEVER include frame numbers in any question or option text.
    Do NOT write "From frame_005", "At frame_008", "As shown in frame_003", etc.
    Describe the scene state in plain language only
    (e.g., "As the person begins lowering toward the seat with their right hand near their face").
    The anchor frame identity is managed by the pipeline - not by the question text.

[WARNING]  NEVER reference events or states that are only visible AFTER the anchor frame.
    The question and all options must be answerable from the anchor frame state alone.
    Do NOT use phrases like "later seen", "as subsequently shown", "which then becomes visible", etc.

==============================
ANSWER KEY CORRECTNESS
==============================

The correct answer must be:
  - Uniquely supported by the physical evidence described in the scene report.
  - Not directly visible in the anchor frame (requires prediction, not recognition).
  - Consistent with the "true_answer_direction" from the skeleton.

Write a clear explanation (2-4 sentences) citing specific scene details (object material,
visible angle, fill level, speed, contact surface, etc.) that support the correct answer.

Write a one-sentence justification for each distractor explaining why it is plausible
but wrong given the specific scene.

==============================
SELECTION - choose the best 3 out of 6
==============================

After completing all 6 QA items, select the best 3 to pass to the next stage.
Prioritise:
  1. Questions that cover different categories (diversity).
  2. Questions with the strongest distractor quality (all 3 wrong options are tempting).
  3. Questions that most clearly require future-process simulation to answer.

Mark the 3 selected items with "selected": true and the other 3 with "selected": false.

==============================
OUTPUT FORMAT
==============================

Return a single JSON object with key "qa_drafts" containing exactly 6 items (ALL text in English):

{
  "qa_drafts": [
    {
      "id": 1,
      "selected": true,
      "category": "C_X_name",
      "category_desc": "Brief English description",
      "question_type": "[ORDER] / [COUNT] / etc.",
      "question": "Full question text referencing specific scene details.",
      "options": {
        "A": "...",
        "B": "...",
        "C": "...",
        "D": "..."
      },
      "answer": "A",
      "explanation": "2-4 sentences citing specific visual evidence for the correct answer.",
      "distractor_justification": {
        "B": "Plausible because X, but wrong because Y.",
        "C": "Plausible because X, but wrong because Y.",
        "D": "Plausible because X, but wrong because Y."
      },
      "difficulty_estimate": "Hard",
      "anchor_frame": "frame_XXX",
      "video_path": "",
      "sample_id": ""
    },
    ...  (6 total, 3 with selected=true)
  ]
}

Output the JSON and stop.
```
\end{tcblisting}

\begin{tcblisting}{
colback=white,
colframe=black,
arc=1mm,
auto outer arc,
title={OpenWorldQA Step 4: Small-model Difficulty Probe},
breakable,
listing only,
listing options={basicstyle=\small\ttfamily, breaklines=true, columns=fullflexible, keepspaces=true, showstringspaces=false}
}
## Agent 4: SmallModelProbe

**Role:** Filters questions by difficulty using a small model (gpt-5-nano). A question is discarded if the small model answers it correctly on both independent attempts; otherwise it passes. The prompt is intentionally minimal.

---

```
Answer the physical reasoning question below using the image provided.
Choose the single best answer. Reply with JSON only: {"answer": "A"}
The answer must be one of: A, B, C, or D.
```

> **Note:** The question text is appended after this prompt in the following format:
> ```
> QUESTION: <question text>
>
> OPTIONS:
>   A: <option A>
>   B: <option B>
>   C: <option C>
>   D: <option D>
> ```
> Each question is tested twice independently with options shuffled randomly to avoid position bias. A question passes only if the small model answers incorrectly on at least one attempt.
\end{tcblisting}

\begin{tcblisting}{
colback=white,
colframe=black,
arc=1mm,
auto outer arc,
title={OpenWorldQA Step 5: Reviewer Prompt},
breakable,
listing only,
listing options={basicstyle=\small\ttfamily, breaklines=true, columns=fullflexible, keepspaces=true, showstringspaces=false}
}
## Agent 5: Reviewer

**Role:** Final quality gate. Receives the anchor frame + context frames + probe-surviving QA items, evaluates each item across five dimensions (answer correctness, anchor validity, distractor plausibility, visual consistency, category alignment), and accepts only items scoring >= 7.

---

```
You are Reviewer - the final quality-gate agent in a multi-stage physical reasoning QA pipeline.

==============================
IMAGE LAYOUT - READ THIS FIRST
==============================
You receive images in two groups, clearly labelled in the INPUT section:

  [ANCHOR]  - exactly 1 image.
              This is the ONLY image the test model sees at evaluation time.
              When you evaluate anchor_validity, evaluate THIS image.

  [CONTEXT] - 0 or more images that follow.
              These are provided for your reference ONLY to verify answer correctness.
              They are NOT available to the test model at evaluation time.

The INPUT section at the bottom of this prompt lists each image in order.
Image 1 in your visual input = the anchor. Images 2+ = context frames.
==============================

PIPELINE CONTEXT:
  - The SmallModelProbe has already rejected questions that gpt-5-nano answered correctly
    on two independent attempts (with shuffled option order).
  - Questions reaching you have PASSED the difficulty gate.
  - Do NOT re-evaluate difficulty. Focus on correctness and plausibility only.

==============================
YOUR REVIEW TASKS (evaluate each QA independently)
==============================

1. ANSWER CORRECTNESS
   Using ALL frames (anchor + context), verify that the claimed correct answer is physically
   accurate. Consider: physical laws, object properties, visible forces, trajectory.
   Reject if the answer key is wrong or ambiguous.

2. ANCHOR VALIDITY
   Look at Image 1 (the anchor frame, filename in INPUT section).
   Does it show sufficient initial-condition cues for a reasoning model to predict the outcome?
   Does it NOT already reveal the answer directly?
   Reject if:
     (a) Image 1 is a blank/cover shot with no relevant physical content, OR
     (b) The outcome is already fully visible in Image 1 (static leakage).

3. DISTRACTOR PHYSICAL PLAUSIBILITY
   For each wrong option independently:
     - Is it physically possible in the real world under some conditions? (MUST be yes)
     - Could a viewer reasonably think it might happen in this specific scene? (MUST be yes)
   Reject immediately if ANY distractor:
     - Defies gravity or levitates without cause
     - Freezes in midair
     - Passes through solid objects
     - Reverses motion spontaneously
     - Is obviously nonsensical in the scene context

4. VISUAL CONSISTENCY
   Does the question text accurately describe what is shown in the frames?
   Reject if the question references objects or actions not visible anywhere in the clip.

5. CATEGORY ALIGNMENT
   Does this QA item genuinely test the assigned category's type of reasoning?
   (Not a blocking reject - note and suggest correction if misaligned.)

==============================
APPROVE if ALL of the following hold:
==============================
  [PASS] Answer key is correct and unambiguous
  [PASS] Image 1 (anchor) provides sufficient initial-condition cues without revealing the answer
  [PASS] Every distractor is physically plausible and locally tempting in this scene
  [PASS] No distractor violates basic physics
  [PASS] Question text matches visible content
  [PASS] Score >= 7

REJECT if ANY of the following hold:
  [REJECT] Answer key is wrong or ambiguous
  [REJECT] Image 1 (anchor) is a cover/blank shot OR already reveals the outcome
  [REJECT] Any distractor defies physics or is obviously nonsensical
  [REJECT] Question references objects or actions not visible in the clip
  [REJECT] Score < 7

==============================
OUTPUT FORMAT
==============================

Return a single JSON object with key "review_list" containing one entry per QA item
(same order as input). ALL text in English.

{
  "review_list": [
    {
      "decision": "accept",
      "score": 8,
      "reasoning": {
        "answer_correctness":       {"pass": true,  "note": "..."},
        "anchor_validity":          {"pass": true,  "note": "..."},
        "distractor_plausibility":  {"pass": true,  "note": "Evaluate each wrong option individually."},
        "visual_consistency":       {"pass": true,  "note": "..."},
        "category_alignment":       {"pass": true,  "note": "..."}
      },
      "suggestions": ""
    },
    ...
  ]
}

Output the JSON and stop.
```
\end{tcblisting}

\end{document}